%% file: main.tex
\documentclass[11pt,letterpaper]{article}

\usepackage[letterpaper,margin=0.98in]{geometry}
\usepackage[utf8]{inputenc}
\usepackage[T1]{fontenc}
\usepackage{microtype}
\usepackage{url}

\usepackage{amsmath}
\usepackage{amssymb}
\usepackage{amsfonts}
\usepackage{amsthm}
\usepackage{mathtools}
\usepackage{nicefrac}

\usepackage{graphicx}
\usepackage{subcaption}
\usepackage{wrapfig}
\usepackage{booktabs}
\usepackage{multirow}
\usepackage{makecell}
\usepackage{float}
\usepackage{threeparttable}

\usepackage[table,xcdraw]{xcolor}
\definecolor{darkgray}{RGB}{50,50,60}
\definecolor{rulegray}{RGB}{180,190,220}
\definecolor{highlight}{gray}{0.92}
\definecolor{lightblue}{RGB}{100,149,237}
\definecolor{sienna}{RGB}{136,45,23}
\definecolor{slategray}{RGB}{90,90,90}
\definecolor{burgundy}{RGB}{128,0,32}
\definecolor{medblue}{RGB}{70,130,180}
\definecolor{ourcolor}{RGB}{235,242,255}
\definecolor{untempcolor}{RGB}{255,240,230}

\definecolor{thmbg}{RGB}{243,245,248}        
\definecolor{thmrule}{RGB}{52,82,115}        
\definecolor{lemmabg}{RGB}{243,245,248}      
\definecolor{lemmarule}{RGB}{75,95,120}      
\definecolor{assbg}{RGB}{243,245,248}        
\definecolor{assrule}{RGB}{105,120,140}      
\definecolor{takeawaybg}{RGB}{243,245,248}   
\definecolor{takeawayrule}{RGB}{90,90,95}    

\usepackage[normalem]{ulem}
\useunder{\uline}{\ul}{}
\usepackage{enumitem}
\usepackage{etoc}

\usepackage{authblk}

\usepackage{tcolorbox}
\tcbuselibrary{skins, breakable}
\usepackage{mdframed}

\newtcolorbox{thmboxinner}{
  enhanced, breakable, rounded corners,
  colback=thmbg, colframe=thmrule,
  boxrule=0pt, leftrule=2.5pt,
  left=10pt, right=8pt, top=4pt, bottom=4pt,
  before skip=8pt, after skip=8pt,
}
\newtcolorbox{lemmaboxinner}{
  enhanced, breakable, rounded corners,
  colback=lemmabg, colframe=lemmarule,
  boxrule=0pt, leftrule=2.5pt,
  left=10pt, right=8pt, top=4pt, bottom=4pt,
  before skip=8pt, after skip=8pt,
}
\newtcolorbox{assumptionboxinner}{
  enhanced, breakable, rounded corners,
  colback=assbg, colframe=assrule,
  boxrule=0pt, leftrule=2.5pt,
  left=10pt, right=8pt, top=4pt, bottom=4pt,
  before skip=8pt, after skip=8pt,
}

\newtcolorbox{takeawaybox}{
  enhanced, breakable, rounded corners,
  colback=takeawaybg, colframe=takeawayrule,
  boxrule=0pt, leftrule=2.5pt,
  left=10pt, right=8pt, top=4pt, bottom=4pt,
  before skip=8pt, after skip=8pt,
  fontupper=\small\itshape,
}

\usepackage{algorithm}
\usepackage{algpseudocode}

\usepackage[textsize=tiny]{todonotes}

\newtheorem{theorem}{Theorem}
\theoremstyle{plain}
\newtheorem{lemma}{Lemma}

\theoremstyle{definition}

\newtheorem{assumption}{Assumption}

\usepackage[
  style=authoryear-comp,
  maxbibnames=15,
  maxcitenames=2,
  giveninits=true,
  natbib=true,
  backend=biber,
  sorting=nyt,
  uniquename=false,
  uniquelist=false,
]{biblatex}
\AtBeginBibliography{\emergencystretch=2em}

\usepackage{hyperref}
\hypersetup{
  colorlinks=true,
  breaklinks=true,
  citecolor=medblue,
  linkcolor=sienna,
  urlcolor=sienna,
  filecolor=black,
  bookmarksopen=false,
}

\DeclareFieldFormat{citehyperref}{%
  \DeclareFieldAlias{bibhyperref}{noformat}%
  \bibhyperref{#1}}
\DeclareFieldFormat{textcitehyperref}{%
  \DeclareFieldAlias{bibhyperref}{noformat}%
  \bibhyperref{%
    #1%
    \ifbool{cbx:parens}
      {\bibcloseparen\global\boolfalse{cbx:parens}}
      {}}}

\savebibmacro{cite}
\savebibmacro{textcite}

\renewbibmacro*{cite}{%
  \printtext[citehyperref]{%
    \restorebibmacro{cite}%
    \usebibmacro{cite}}}

\renewbibmacro*{textcite}{%
  \ifboolexpr{
    ( not test {\iffieldundef{prenote}}
      and test {\ifnumequal{\value{citecount}}{1}} )
    or
    ( not test {\iffieldundef{postnote}}
      and test {\ifnumequal{\value{citecount}}{\value{citetotal}}} )
  }
    {\DeclareFieldAlias{textcitehyperref}{noformat}}
    {}%
  \printtext[textcitehyperref]{%
    \restorebibmacro{textcite}%
    \usebibmacro{textcite}}}
\usepackage[capitalize,noabbrev]{cleveref}
\renewcommand{\citet}[1]{\citeauthor{#1}~(\citeyear{#1})}

\crefname{theorem}{Theorem}{Theorems}        \Crefname{theorem}{Theorem}{Theorems}
\crefname{lemma}{Lemma}{Lemmas}              \Crefname{lemma}{Lemma}{Lemmas}
\crefname{proposition}{Proposition}{Propositions}
\Crefname{proposition}{Proposition}{Propositions}
\crefname{corollary}{Corollary}{Corollaries} \Crefname{corollary}{Corollary}{Corollaries}
\crefname{definition}{Definition}{Definitions}
\Crefname{definition}{Definition}{Definitions}
\crefname{assumption}{Assumption}{Assumptions}
\Crefname{assumption}{Assumption}{Assumptions}
\crefname{remark}{Remark}{Remarks}           \Crefname{remark}{Remark}{Remarks}
\crefname{example}{Example}{Examples}        \Crefname{example}{Example}{Examples}
\crefname{algorithm}{Algorithm}{Algorithms}  \Crefname{algorithm}{Algorithm}{Algorithms}

\usepackage{amsmath}
\usepackage{setspace}
\usepackage{enumitem}
\usepackage{pifont}

\newcommand{\ours}[0]{TTS}
\newcommand{\ignore}[1]{}

\newcommand{\BlueComment}[1]{\hfill{\color{blue}// #1}}

\definecolor{ForestGreen}{RGB}{34, 139, 34}
\newcommand{\cmark}{\textcolor{ForestGreen}{\ding{51}}}
\newcommand{\xmark}{\textcolor{red}{\ding{55}}}

\usepackage{titlesec}
\titlespacing*{\section}{0pt}{0.6\baselineskip}{0pt}
\titlespacing*{\subsection}{0pt}{0.3\baselineskip}{-0.2\baselineskip}

\title{Test-Time Speculation \vspace{3mm}
\footnotetext{Corresponding author: \texttt{avinkumar@utexas.edu}. Project page: \url{https://avinkumarut.github.io/TTS/}}}
\author{Avinash Kumar}
\author{Sujay Sanghavi}
\author{Poulami Das}
\affil{The University of Texas at Austin}
\date{}

\begin{document}

\etocdepthtag.toc{mainpaper}
\maketitle

\input{Sections/0_Abstract}
\input{Sections/1_Introduction}
\input{Sections/2_Background}
\input{Sections/3_Design}
\input{Sections/4_Experiments}
\input{Sections/5_Conclusion}
\input{Sections/6_Acknowledgments}

\printbibliography

\clearpage
\input{Sections/7_Appendix}

\end{document}

%% file: Sections/0_Abstract.tex
\begin{abstract}

Speculative decoding accelerates LLM inference by using a fast draft model to generate tokens and a more accurate target model to verify them. Its performance depends on the \textit{acceptance length}, or number of draft tokens accepted by the target. Our studies show that the acceptance length of even state-of-the-art speculators, like DFlash, EAGLE-3, and PARD, degrades with generation length, reaching values close to 1 (i.e., no speedup) within a few thousand output tokens, making speculators ineffective for long-response tasks. Acceptance lengths decline because most speculators are trained offline on short sequences, but are forced to match the target on much longer outputs at inference, well beyond their training distribution.

To address this issue, we propose \textit{Test-Time Speculation~(\ours{})}, an online distillation approach that continuously adapts the speculator at test time. \ours{} leverages the key insight that the token verification step already invokes the target model for each draft token, providing the training signal needed to adapt the draft at no additional cost. Treating the draft as the student and the target as the teacher,~\ours{} adjusts the draft over several speculation rounds, with each update improving the draft's accuracy as generation proceeds. Our results across ten models from the Qwen-3, Qwen-3.5, DeepSeek-R1, GPT-OSS, MiniMax, and Llama3.1 families show that~\ours{} improves acceptance lengths over state-of-the-art speculators by up to $72\%$ and $41\%$ on average, with the benefits scaling with generation lengths.

\end{abstract}

%% file: Sections/1_Introduction.tex
\section{Introduction}

Large Language Models (LLMs) can perform various tasks, such as summarization~(\cite{stiennon2020learning}), code generation~(\cite{chen2021evaluating}), content creation~(\cite{brown2020language}), etc., and are becoming increasingly ubiquitous. However, they pose critical performance concerns because tokens can only be generated auto-regressively. Speculative decoding~(\cite{leviathan2023fast}) is a widely adopted inference technique that breaks this dependency by using a fast but less accurate draft model to quickly generate a sequence of tokens and an accurate but slow target model to verify these tokens in parallel. The draft model optimizes for low latency, whereas the target model \textit{ensures} that the final output tokens \textit{exactly match} those that it would have produced auto-regressively. The performance of speculative decoding depends on the number of draft tokens accepted by the target model per round, also known as the \textit{acceptance length}. In practice, \textit{higher acceptance lengths are desirable}.

\begin{figure}[t]
    \centering
    \includegraphics[width=1.0\textwidth]{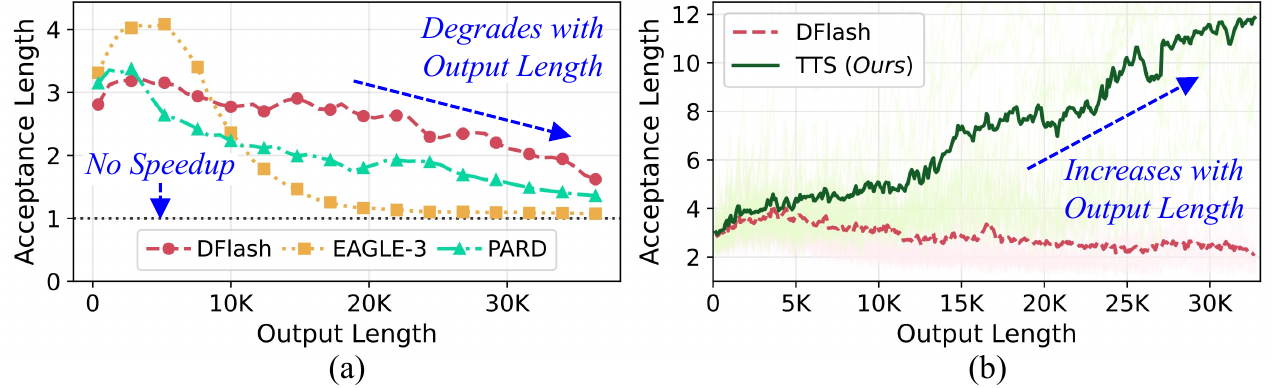}
    \caption{(a) Acceptance length (AL) for the LiveCodeBench dataset on Qwen3-8B with increasing output lengths for three speculators. Beyond a certain number of output tokens, the draft model fails to accurately approximate the target model and the AL becomes vanishingly low (too close to 1), yielding practically no benefits. (b) Our proposal, \textit{Test-Time Speculation (\ours{})}, adapts the draft model in real time to address this limitation and consistently improves acceptance lengths. }
    \label{fig:intro}
\end{figure}

State-of-the-art speculators improve acceptance lengths by increasing the accuracy of the draft model. EAGLE-3~(\cite{li2025eagle}) produces draft tokens using a transformer layer that conditions on multi-level hidden states from the target model. PARD~(\cite{an2025pard}) appends mask-token placeholders to the current sequence and fine-tunes a pretrained auto-regressive draft model to predict all masked positions in parallel. DFlash~(\cite{chen2026dflash}) uses a block-diffusion drafter and injects fused multi-layer target features into the speculator's KV cache. However, although effective for tasks producing only a small number of output tokens, these methods do not retain their benefits for \textit{long-response tasks} producing thousands of tokens. For example, Figure~\ref{fig:intro}(a) shows the average acceptance length with increasing number of output tokens for LiveCodeBench~(\cite{jain2024livecodebench}) on Qwen3-8B using the above speculators. The acceptance length decreases with output length. More importantly, after a certain point, speculative decoding yields diminishing returns because acceptance lengths are too low to offer any practical speedup benefits. For instance, the acceptance length reduces to 1.1 beyond ${\sim}20K$ tokens for EAGLE-3, a widely used speculator in industrial serving systems~(\cite{tang2025efficient}), defeating the purpose of speculative decoding at that point. As long-response tasks dominate many real-world LLM deployment scenarios~(\cite{jain2024livecodebench,rein2023gpqa}), improving acceptance lengths is critical to retain the benefits of speculative decoding.

The degradation in acceptance length stems from a mismatch between the draft's training and inference distributions. State-of-the-art speculators are small networks trained on short sequences~(\cite{chen2026dflash, li2025eagle}), typically less than $2K$ tokens. However, at inference time, the draft is forced to match the target on much longer sequences, causing its predictions to progressively diverge from the target model. Consequently, a larger number of draft tokens become inaccurate and are eventually rejected by the target model, decreasing acceptance lengths. Closing this gap requires adapting the draft model to match the target distribution at inference time.

\textbf{Our Proposal:} We propose \textit{\underline{T}est-\underline{T}ime \underline{S}peculation (\ours{})}, an online distillation approach that adapts the draft model to match the target's distribution at inference time.~\ours{} exploits the key insight that each token verification step already produces the supervision needed to adapt the draft, as the target model computes its full distribution over all generated draft tokens.~\ours{} leverages this signal by treating the target as the teacher, the draft as the student, and the target's distribution over the draft tokens as the distillation sample on which the draft performs an optimization step. Our evaluations across ten state-of-the-art models show that~\ours{} consistently improves acceptance lengths by up to $72\%$ and $41\%$ on average, with the benefits scaling with generation lengths, as shown in Figure~\ref{fig:intro}(b).

Overall, this paper makes the following contributions:
\begin{enumerate}[leftmargin=8pt, itemindent=5pt, labelsep=5pt,itemsep=5pt]
    \item We observe that the efficacy of state-of-the-art speculators, like DFlash, EAGLE-3, and PARD, degrades with generation length, with acceptance lengths decreasing to values too close to 1 within just a few thousand tokens, making them ineffective for real-world long-response tasks.
    
    \item We demonstrate that acceptance length degrades because draft models are trained on short sequences, but are forced to match the target distribution on much longer sequences at inference. 
    
    \item We propose \textit{\underline{T}est-\underline{T}ime \underline{S}peculation (\ours{})}, an online distillation approach that leverages the insight that the token verification step already produces the supervision signal needed to adapt the draft model at inference time. We enable this by using the target as the teacher, the draft as the student, and the target's distribution over the drafted tokens in each round as distillation samples.
    
    \item Our experiments across three state-of-the-art speculators show that~\ours{} improves acceptance length by up to~$72\%$ and $41\%$ on average, with benefits scaling with increased generation lengths.
\end{enumerate}

%% file: Sections/2_Background.tex
\section{Problem: Acceptance Length Decreases With Output Length}

\begin{figure*}[htp]
\begin{center}
    \centering
    \includegraphics[width=1.0\textwidth]{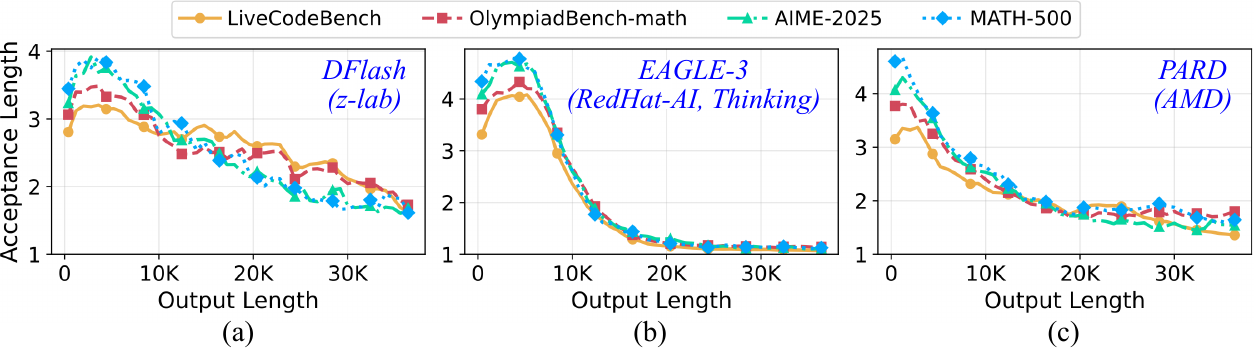}
    \caption{Acceptance length of four tasks using (a) DFlash, (b) EAGLE-3, and (c) PARD speculators on the Qwen3-8B model. Acceptance length consistently decreases with increasing generation length.}
\label{fig:al_bottleneck}
\end{center}
\end{figure*}

Although effective for tasks generating short responses, such as GSM8K~(\cite{gsm8k}) and HumanEval~(\cite{humaneval}), our experiments show that state-of-the-art speculators are ineffective for long-response generation tasks producing thousands of output tokens. Long-response tasks, such as content creation~(\cite{longwriter}), long-form reasoning~(\cite{guo2025deepseek}), and agentic workflows~(\cite{yao2023react}), are critical in modern real-world LLM deployment scenarios and continue to dominate most inference serving systems~(\cite{qin2024mooncake, agrawal2024taming}). However, our experiments show that acceptance length decreases with increasing number of output tokens even with state-of-the-art speculators. Figure~\ref{fig:al_bottleneck}(a-c) shows the mean acceptance length across 64 sequences for four different tasks (LiveCodeBench, OlympiadBench-Math, AIME 2025, MATH-500) on the Qwen3-8B model using three state-of-the-art speculators (DFlash, EAGLE-3 and PARD). We observe that acceptance lengths decrease consistently for all tasks and speculators. For the MATH-500 dataset, the average acceptance length is $3.7$ for the first $10K$ output tokens, but sharply decreases to $1.5$ for the last $10K$ output tokens in the sequence. In fact, when EAGLE-3 is used as the speculator, acceptance length reduces to $1.1$ for all the tasks after generating ${\sim}20K$ tokens and therefore, practically yields no speedup.

\begin{figure*}[htp]
\begin{center}
    \centering
    \includegraphics[width=1.0\textwidth]{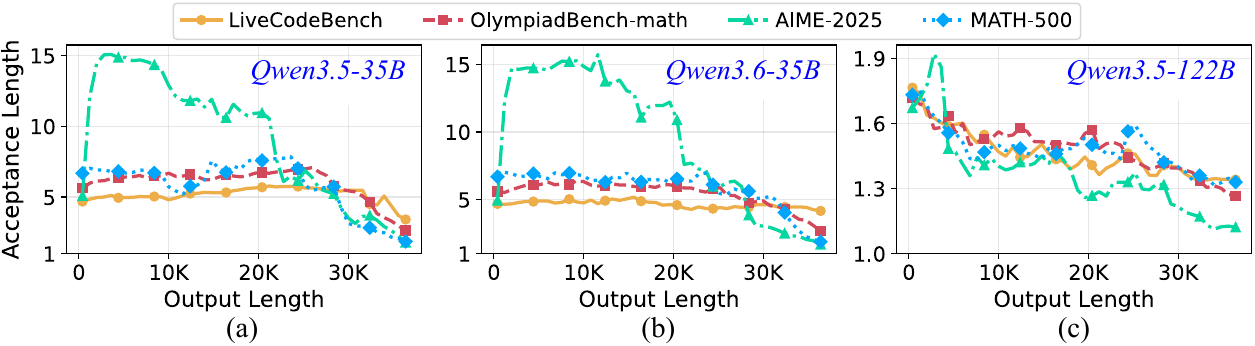}
    \caption{Acceptance length of four tasks using the DFlash speculator on (a) Qwen3.5-35B, (b) Qwen3.6-35B, and (c) Qwen3.5-122B models. Acceptance length consistently reduces with increasing output generation lengths even when larger models are used.}
\label{fig:al_bottleneck_b}
\end{center}
\end{figure*}

Moreover, the problem persists even when larger models are used. Figure~\ref{fig:al_bottleneck_b} shows that acceptance length consistently degrades for the same four tasks even when using the most recent DFlash speculator when larger models, such as Qwen3.5-35B, Qwen3.6-35B, and Qwen3.5-122B, are used. The acceptance length decreases from $15$ to $1.7$ for the AIME 2025 task on the Qwen3.6-35B model within a span of $25K$ tokens, with a sharp decline from $11$ to $3.8$ going from $20K$ to $28K$ tokens.

Higher acceptance lengths are necessary to attain meaningful performance benefits with speculative decoding. Acceptance length, in turn, depends on the accuracy of the draft model. We therefore diagnose why the draft model degrades as the generation length increases.

\section{Diagnosis: Why The Draft Degrades At Higher Output Lengths}
\label{sec:diagnosis}

The draft is a small model trained offline to match the target model's token distribution. Training is typically performed on short sequences ${\leq}2K$ tokens~(\cite{chen2026dflash, li2025eagle}) drawn from instruction-following datasets such as ShareGPT~(\cite{sharegpt}) and UltraChat~(\cite{ultrachat}). However, at inference time, the draft is forced to match the target on much longer sequences, causing its predictions to progressively diverge from the target and acceptance length to degrade.

\begin{figure*}[htp]
\begin{center}
    \centering
    \includegraphics[width=1.0\textwidth]{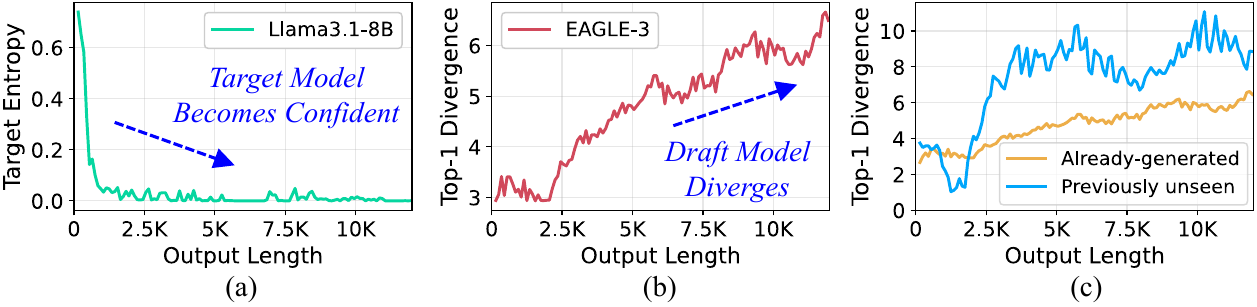}
    \caption{Distribution entropy (in nats) for Llama3.1-8B (target) with EAGLE-3 (draft). (a) Target entropy decreases with output length, indicating that the target becomes increasingly confident in its predictions. (b) Over the same range, the draft's cross-entropy on the target's $Top{-}1$ token increases sharply. (c) This divergence persists regardless of whether the token appeared earlier or was unseen.}
\label{fig:diagnosis}
\end{center}
\end{figure*}

Figure~\ref{fig:diagnosis}(a) shows that the distribution entropy $H(p^i_\text{target})$ of the target model decreases with output length, indicating that the target model becomes increasingly confident in its predictions. Therefore, with each additional token, the output of the target model becomes more predictable. In contrast, Figure~\ref{fig:diagnosis}(b) shows that the cross-entropy of the draft on the target's $Top{-}1$ token ($-\log p^i_\text{draft}(t^*_i)$) increases monotonically from $2.8$ to $6.4$ nats over the same range. At a generation length of $12K$ tokens, the draft assigns a negligible probability of $1/600$ to the target's $Top{-}1$ token. Thus, the draft's generated tokens \textit{drift} farther from the target with output length, even though the target model itself becomes more decisive. Furthermore, Figure~\ref{fig:diagnosis}(c) shows that cross-entropy increases steeply irrespective of whether the token ID was previously unseen or had already been generated in the sequence. To summarize, the draft loses agreement with the target because its offline training on short sequences leaves it miscalibrated with the target at higher output lengths. \textit{Closing this gap requires adapting the draft model to the target's trajectory at inference time.}

%% file: Sections/3_Design.tex
\section{Our Proposal: Test-Time Speculation (\ours{})}
\label{sec:design}

The core algorithmic realization of our paper is that the existing speculative decoding workflow already has all the ingredients needed for online on-policy distillation,~\textit{if} one were willing to interleave generation with updates to the speculator. We now develop notation to formalize this intuition.

{\bf Setup:} Let $p(x)$ denote the target model, i.e., the large base model that we want to generate with. Typically, this is a large auto-regressive model, and vanilla text generation from this model involves iteratively sampling from $p(\cdot)$: 
\[
x_{t+1} ~ \sim ~ p( X_{t+1} \, | \, x_{1:t} \,) \quad \text{(Vanilla auto-regressive generation)}
\]
Here, we use the standard notation $x_{1:t}$ to denote the sequence $x_1,\ldots,x_t$. Each output token, therefore, requires an entire forward pass through the target model $p(\cdot)$, a cost that grows with model size and becomes prohibitive for state-of-the-art LLMs.

Standard speculative decoding pairs the target with a much smaller, and therefore faster, auxiliary model $q(\cdot)$, called the \textit{draft}. The draft model (or speculator) itself need not be auto-regressive. The most capable modern speculators, including the EAGLE-3 and DFlash drafters considered in this work, also reuse the target model's intermediate hidden states. Token generation proceeds in rounds. In each round, the draft model first generates a canvas of tokens. Next, the target model verifies this entire canvas in parallel and ``accepts'' a prefix of the draft tokens. The next round begins from the last accepted token, and since the target model has already executed a forward pass over the entire canvas, its hidden states are consistent up to this new starting point.

\textbf{Standard speculative decoding:} Suppose that we are currently at time step $t$, i.e., tokens $x_{1:t}$ have been verified by the target model, and we want to speculate-verify the next few tokens.
\begin{enumerate}
    \item First, a canvas of tokens is generated from $q(\cdot)$, and then this canvas is evaluated by the target model. If the canvas length is $C$,
    \[
    \text{Generate} \quad  x_{t+1\,:\, t+C} ~ \sim ~ q \left ( X_{t+1\,:\, t+C} \, | \, x_{1\,:\,t} \right ) \quad \text{(Canvas tokens from drafter)}
    \]
    \item Recall that the target model has already processed $x_{1:t}$. It now appends the canvas tokens $x_{t+1},  \ldots , x_{t+C}$ to this, and executes a single forward pass. This yields all the sequentially conditioned likelihoods for canvas tokens:
    \[
    \text{Evaluate} ~~ p(x_{t+k}|x_{1\,:\,t+k-1}) ~ \text{for all tokens} ~ 1\leq k \leq C ~~ \text{(All in one forward pass)}
    \]
    Then, canvas tokens are considered from left to right for acceptance, as follows: 
    \begin{algorithmic}
        \State $k \gets 1$
        \While{$k \leq C$}
            \State accept $x_{t+k}$ with probability $\min \left \{ \, 1 \, , \, \dfrac{p(x_{t+k}|x_{1\,:\,t+k-1})}{q(x_{t+k}|x_{1\,:\,t+k-1})} \right \}$
            \If{$x_{t+k}$ rejected} \textbf{break} \EndIf
            \State $k \gets k+1$
        \EndWhile
    \end{algorithmic}
        
    \item Advance $t$ to be the last accepted token, and go back to step 1.
\end{enumerate}

The key metric of interest in speculative decoding is the~\textbf{acceptance length}, i.e., the average number of tokens accepted in each round. The higher the acceptance length, the greater the speedup, as each round has the same computational cost regardless of how many tokens are accepted per round.

\subsection{Test-Time Speculation}

In the standard workflow above, each time a draft canvas is produced, we naturally have access to both the speculator's probabilities $q(x_{t+1}, \ldots , x_{t+C} | x_1, \ldots, x_t)$ and the target's probabilities $p(x_{t+1}, \ldots , x_{t+C} | x_1, \ldots, x_t)$. Our~\textbf{main insight} is to treat the draft model as the student, the target model as the teacher, and the generated-and-evaluated canvas as a distillation-style training sample for the student. After each round of speculation and verification, our idea is to update the draft model via a single gradient step on a distillation-style loss---before the next canvas is generated. In practice, the draft need not be updated after every round; the update can be applied lazily every few rounds and pipelined on a separate CUDA stream. We study the impact of these choices in Section~\ref{sec:results}.

Formally, suppose as before that generation is at time $t$. Since the speculator is updated over time as generation progresses, we denote it by $q_t(\cdot)$ instead of just $q(\cdot)$. Initially, $q_t(\cdot)$ is set to the existing draft, i.e., $q_0 = q$. We now explain the update mechanism in detail:

First, we follow the standard workflow and~\textit{(a)} use this speculator $q_t(\cdot)$ to generate the canvas $x_{t+1:t+C}$,~\textit{(b)} evaluate the target model likelihoods on this canvas, and~\textit{(c)} determine the acceptance length $\tau$.
We now update $q_t(\cdot)$ to obtain the new speculator $q_{t+\tau}(\cdot)$. This is done via \textbf{one optimization step on the standard drafter distillation loss} on the canvas. Specifically, let $x_\mathcal{C} = x_{t+1},\ldots,x_{t+C}$ denote the sequence of canvas tokens; the update then minimizes the objective:
\begin{equation}
\label{eq:tts-loss}
L_t({\color{blue}q}) ~ = ~ \widetilde{\mathrm{KL}}\left ( \, p(x_\mathcal{C}|x_{1:t})\,||\,{\color{blue}q}(x_\mathcal{C}|x_{1:t})\,\right ) ~ + ~ \lambda \, \widetilde{\mathrm{KL}} \left ( \, q_t(x_\mathcal{C}|x_{1:t})\,||\,{\color{blue}q}(x_\mathcal{C}|x_{1:t})\,\right )
\end{equation}

The first term is an approximate KL divergence (specified below) between the target $p$ and the new $q$, while the second is a regularization term that prevents $q$ from drifting too far from $q_t$. Here
\[
\widetilde{\mathrm{KL}}\left ( \, p(x_\mathcal{C}|x_{1:t})\,||\,{\color{blue}q}(x_\mathcal{C}|x_{1:t})\,\right ) ~ = ~ \sum_{k = 1}^C
 \, w_k \, \mathrm{KL}(p(X_{t+k}|x_{1:t+k-1})||q(X_{t+k}|x_{1:t+k-1}))
 \]
 and $\mathrm{KL}\, \left ( \, p(X_{t+k}|x_{1:t+k-1}) \, || \, q(X_{t+k}|x_{1:t+k-1})\, \right )$ is the standard KL divergence between the target and draft distributions at position $t+k$, conditioned on the prefix $x_{1:t+k-1}$. With uniform weights ($w_k{=}1$ for all $k$), this reduces to the standard distillation loss, with $p(\cdot|x_{1:t})$ as the teacher, $q(\cdot|x_{1:t})$ as the student, and $x_{\mathcal{C}}$ as the string on which the loss is calculated.
 
Finally, the $w_k$ are position-dependent weights that decrease with $k$, as a rejected token invalidates all succeeding canvas tokens, making early positions more important. We reuse these weights directly from the training codebase of each speculator. Algorithm~\ref{alg:tts} gives an overview of the~\ours{} algorithm.

\begin{algorithm}[H]
\caption{Test-Time Speculation (TTS)}
\label{alg:tts}
\begin{spacing}{1.1}
\begin{algorithmic}[1]
    \State \textbf{Input:} Target model $p$, draft model $q$, learning rate $\eta$, regularizer $\lambda$, canvas size $C$
    \State \hspace{1.3em} Input prompt $x_{1:T_0}$ \BlueComment{Prompt of length $T_0$ tokens}
    \State \hspace{1.3em} Max new tokens $L_{\max}$ \BlueComment{Generation length cap}
    \State \hspace{1.3em} Weights $\{w_k\}_{k=1}^{C}$ \BlueComment{Per-position weights, matching the drafter's training recipe}
    \State \textbf{Algorithm:}
    \State $t \gets T_0$,\ \ $q_t \gets q$
    \While{$x_t \ne \textsc{eos}$ \textbf{and} $t - T_0 < L_{\max}$}
        \State \textbf{Draft:}\ \ sample canvas $x_{t+1:t+C} \sim q_t(\,\cdot\, \mid x_{1:t})$
        \State \textbf{Verify:}\ \ compute $p(x_{t+k}\mid x_{1:t+k-1})$ for $k=1,\ldots,C$ in one forward pass of $p$
        \State \textbf{Accept:}\ \ run standard speculative rejection sampling; let $\tau$ be the number of accepted tokens
        \State \textbf{Loss:}\ \ $L_t({\color{blue}q}) \gets \widetilde{\mathrm{KL}}(p \,\|\, {\color{blue}q}) + \lambda\,\widetilde{\mathrm{KL}}(q_t \,\|\, {\color{blue}q})$ \BlueComment{Distillation loss + regularization}
        \Statex \hspace{1.5em}\textit{where}\ \ $\widetilde{\mathrm{KL}}(a\|b) := \sum_{k=1}^{C} w_k\,\mathrm{KL}\!\bigl(a(X_{t+k}\mid x_{1:t+k-1}) \,\big\|\, b(X_{t+k}\mid x_{1:t+k-1})\bigr)$
        \State \textbf{Update:}\ \ $q_{t+\tau} \gets q_t - \eta\,\nabla_{\color{blue}q} L_t({\color{blue}q})\big\rvert_{{\color{blue}q}=q_t}$ \BlueComment{One gradient step on $L_t$}
        \State $t \gets t + \tau$
    \EndWhile
    \State \textbf{return} $x_{1:t}$
\end{algorithmic}
\end{spacing}
\end{algorithm}

For EAGLE-3, we instead use the~\textit{LK loss}, following prior work~(\cite{samarin2026lk}) and the SpecForge~(\cite{li2026specforge}) training recipe. Concretely, $\widetilde{\mathrm{KL}}$ in~\eqref{eq:tts-loss} is replaced with:
\[
  \widetilde{\mathrm{LK}}\left ( \, p(x_\mathcal{C}|x_{1:t})\,||\,{\color{blue}q}(x_\mathcal{C}|x_{1:t})\,\right )
  = - \sum_{k = 1}^C w_k \log \sum_{x_{t+k}} \min \left\{ p(x_{t+k}|x_{1:t+k-1}) , {\color{blue}q}(x_{t+k}|x_{1:t+k-1}) \right\}
\]
The overall loss retains the same form: a $w_k$-weighted sum over the $C$ canvas positions.

%% file: Sections/4_Experiments.tex
\section{Evaluation Methodology}
\label{sec:methodology}

\noindent \textbf{Benchmark:} We consider eight single-turn benchmarks, spanning competition mathematics (AIME 2024/2025, MATH-500, and OlympiadBench-Math), physics (OlympiadBench-Physics), graduate-level science (GPQA $\diamond$), theorem reasoning (TheoremQA), and coding (LiveCodeBench), all of which elicit output lengths ranging up to $32K$ tokens. We also consider multi-turn agentic tasks drawn from OpenThoughts~(\cite{openthoughts-agent}). We provide more details about these tasks in Appendix~\ref{app:benchmarks}.

\noindent \textbf{Setup:} We conduct all our experiments on a \textit{commercial} Amazon EC2 G7e instance with 8$\times$NVIDIA RTX PRO 6000~\textit{Blackwell} GPUs, each with 96GB memory. We implement~\ours{} on top of the public EAGLE-3~(\cite{li2025eagle}), DFlash~(\cite{chen2026dflash}), and PARD~(\cite{an2025pard}) codebases.

\noindent \textbf{Models and Baselines:} We evaluate on ten state-of-the-art models spanning the Qwen-3~(\cite{yang2025qwen3}), Qwen-3.5~(\cite{qwen3.5}), DeepSeek-R1-Distill-Qwen~(\cite{guo2025deepseek}), GPT-OSS~(\cite{openai2025gptoss120bgptoss20bmodel}), MiniMax~(\cite{chen2026minimax}) and Llama3.1~(\cite{grattafiori2024llama}) families. We apply~\ours{} to, and compare against, the official EAGLE-3, DFlash, and PARD checkpoints. Please refer to Appendix~\ref{app:models} for a detailed description of baselines and the exact HuggingFace identifiers.

\noindent \textbf{Metrics:} We evaluate~\ours{} on mean acceptance length and \textit{throughput}. Acceptance length measures how well the speculator matches the target distribution, and a higher value directly translates into fewer invocations of the target model. It is \textit{hardware-independent}, decoupling speculator quality from the underlying software stack. Throughput, in contrast, depends on the inference stack and is sensitive to system-level optimizations like fused kernels and quantization. We do not report accuracy metrics as they \textit{remain unchanged} as~\ours{} only adapts the draft model and leaves the target untouched. Thus, output tokens remain consistent with the target model~(\cite{leviathan2023fast}).

Our comprehensive selection of benchmarks, models, and comparison against state-of-the-art speculators enables a rigorous evaluation of~\ours{} across diverse real-world tasks and model sizes.

\section{Experimental Results}
\label{sec:results}

\subsection{\ours{} Is Generalizable Across Speculators, Models And Tasks}
Table~\ref{tab:dflash-comp} shows the performance of~\ours{} against DFlash. We observe that \ours{} consistently outperforms vanilla DFlash across all benchmarks and models, improving AL by up to~$72\%$ and $41\%$ on average. We show that~\ours{} generalizes across both EAGLE-3 and PARD in Appendix~\ref{app:spec-results}.

\begin{table}[!htb]
    \caption{Acceptance length of~\ours{} and DFlash across models and tasks. $\Delta(\%)$ denotes the relative improvement of~\ours{} over DFlash.~\ours{} consistently achieves higher acceptance length.}
    \label{tab:dflash-comp}
    \centering
    \begin{tabular}{l @{\hskip 0em} c@{\hskip 3pt}c@{\hskip 3pt}c @{\hskip 0.5em} c@{\hskip 3pt}c@{\hskip 3pt}c @{\hskip 0.5em} c@{\hskip 3pt}c@{\hskip 3pt}c}
    \toprule
    \multirow{2}{*}{Dataset} 
    & \multicolumn{3}{c}{Qwen/Qwen3-4B} 
    & \multicolumn{3}{c}{Qwen/Qwen3-8B} 
    & \multicolumn{3}{c}{Qwen/Qwen3.5-122B} \\
    \cmidrule(lr){2-4} \cmidrule(lr){5-7} \cmidrule(lr){8-10}
    & DFlash & TTS~(\textit{Ours}) & $\Delta(\%)$ & DFlash & TTS~(\textit{Ours}) & $\Delta(\%)$ & DFlash & TTS~(\textit{Ours}) & $\Delta(\%)$ \\
    \midrule
    AIME 2024       & 3.3 & {\bf 4.6} & 38.0 & 3.3 & {\bf 4.6} & 38.5 & 2.9 & \textbf{4.8} & 62.2 \\
    AIME 2025       & 3.2 & {\bf 4.7} & 47.0 & 3.1 & {\bf 4.2} & 33.1 & 2.9 & \textbf{4.9} & 71.8 \\
    GPQA $\diamond$ & 3.6 & {\bf 5.2} & 47.2 & 3.2 & {\bf 4.7} & 47.3 & 2.6 & \textbf{3.5} & 33.9 \\
    LiveCodeBench   & 3.0 & {\bf 4.7} & 55.2 & 2.9 & {\bf 4.5} & 55.2 & 2.4 & \textbf{4.0} & 63.5 \\
    MATH-500        & 3.7 & {\bf 4.3} & 13.8 & 3.8 & {\bf 4.3} & 13.3 & 3.4 & \textbf{4.5} & 32.2 \\
    OlyBench-Math   & 3.0 & {\bf 4.4} & 48.9 & 2.7 & {\bf 4.1} & 50.3 & 3.4 & \textbf{4.7} & 40.2 \\
    OlyBench-Phy    & 3.1 & {\bf 4.5} & 48.1 & 2.8 & {\bf 4.3} & 53.1 & 3.0 & \textbf{3.9} & 28.4 \\
    TheoremQA       & 3.6 & {\bf 4.4} & 23.4 & 3.5 & {\bf 4.3} & 22.8 & 3.3 & \textbf{4.2} & 25.1 \\
    \bottomrule
    \end{tabular}
\end{table}

\subsection{Improvement Scales With Generation Length}

\begin{figure*}[htp]
\begin{center}
    \centering
    \includegraphics[width=1.0\textwidth]{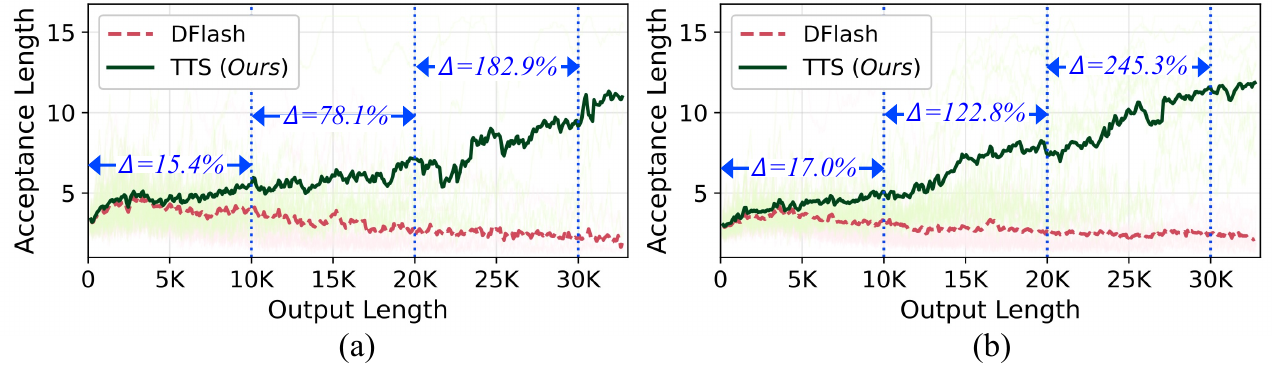}
    \caption{Acceptance length of~\ours{} versus DFlash for (a) AIME 2024 and (b) LiveCodeBench on Qwen3-8B. The improvement ($\%$) increases sharply with output length as~\ours{} refines the drafter.}
\label{fig:increasing-gains-tts}
\end{center}
\end{figure*}

\ours{} accumulates training signal across speculation rounds. Consequently, sequences with longer outputs provide more opportunity to refine the draft model and improve its accuracy further as generation lengths increase. Figure~\ref{fig:increasing-gains-tts} shows that the improvement in acceptance length with~\ours{} increases sharply with generation (or output) length. On AIME 2024, it only improves by $15\%$ for the first $10K$ tokens, but the gap widens to $183\%$ for the tokens in the $20{-}30K$ range. Similarly, on LiveCodeBench, the acceptance length only improves by $17\%$ for the first $10K$ tokens but 
goes up to $245\%$ for the tokens in the $20{-}30K$ range. With the increasing dominance of tasks requiring LLMs to produce long output responses, \ours{} provides a critical advantage in those regimes. 

\subsection{Beyond Single-Turn Reasoning: \ours{} In Agentic Workloads}
\label{sec:agentic}

Unlike the single-turn workloads considered so far, agentic workloads unfold over many~\textit{turns}. In each turn, the model typically emits an~\textit{action}, i.e., a call to an external tool, and the environment returns an~\textit{observation}, i.e., the output produced by the tool. Both are appended to the context for the following turn, and this repeats until the request is served. Such workloads are a particularly favorable setting for~\ours{}: token generation continues across the entire session, so the speculator keeps accumulating training signal instead of resetting after a single response. Note that only the model's output tokens are speculated and verified, and thus only these are used to adapt the speculator.

\begin{figure*}[htp]
\begin{center}
    \centering
    \includegraphics[width=1.0\textwidth]{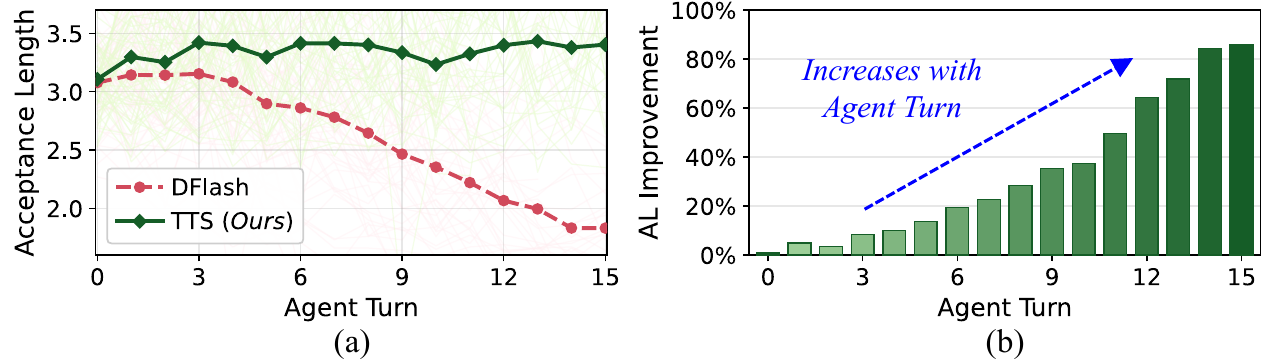}
    \caption{Acceptance length (AL) of~\ours{} versus DFlash on multi-turn agentic tasks from OpenThoughts with MiniMax-M2.5 as the target. (a) The AL of DFlash degrades as the session progresses, whereas~\ours{} sustains it across turns. (b) The improvement ($\%$) of~\ours{} over DFlash grows with turns, reaching $86\%$ at turn $15$, as~\ours{} keeps adapting the draft throughout the session.}
\label{fig:agent}
\end{center}
\end{figure*}

Figure~\ref{fig:agent} shows the improvement in AL over turns on tasks drawn from OpenThoughts-Agent-SFT-100K~(\cite{openthoughts-agent}), orchestrated by the Terminus-2 agent protocol~(\cite{merrill2026terminal}) with MiniMax-M2.5~(\cite{chen2026minimax}) as the target. We observe that the improvement in AL grows monotonically with turns, from $3.5\%$ at turn $2$ to $86.0\%$ at turn $15$. While the AL of the baseline DFlash falls from $3.1$ to $1.8$ over the session,~\ours{} sustains an AL between $3.3$ and $3.4$ throughout.

\subsection{Extensive Offline Training Does Not Eliminate The Need For \ours{}}
\label{sec:extensive}

\ours{} improves AL even on speculators trained extensively on long-context data. To demonstrate this, we evaluate~\ours{} on the GPT-OSS-120B model using two vendor-released EAGLE-3 checkpoints. Both checkpoints are extensively trained on long-context data, with traces extending well beyond $100K$ output tokens. Table~\ref{tab:lc-gpt} shows that~\ours{} improves AL by $39.0\%$ and $44.3\%$ respectively. Thus, extensive offline training alone does not eliminate the need for online adaptation with~\ours{}.

\begin{table}[!htb]
    \caption{Acceptance length of EAGLE-3 and~\ours{} on GPT-OSS-120B for two vendor-released speculators that are extensively trained on long-context data. $\Delta(\%)$ denotes the relative improvement of~\ours{} over EAGLE-3.~\ours{} improves AL across all benchmarks and both speculators.\vspace{0.05em}}
    \label{tab:lc-gpt}
    \centering
        \begin{tabular}{lcccccc}
        \toprule
        \multirow{2}{*}{Dataset}
        & \multicolumn{3}{c}{nvidia/gpt-oss-120b-Eagle3-v3}
        & \multicolumn{3}{c}{nvidia/gpt-oss-120b-Eagle3-long-context} \\
        \cmidrule(lr){2-4} \cmidrule(lr){5-7}
        & EAGLE-3 & TTS~(\textit{Ours}) & $\Delta(\%)$ & EAGLE-3 & TTS~(\textit{Ours}) & $\Delta(\%)$ \\
        \midrule
        AIME 2024          & 4.9 & \textbf{6.4} & 28.7 & 4.0 & \textbf{5.9} & 47.1 \\
        AIME 2025          & 4.6 & \textbf{6.2} & 34.1 & 3.9 & \textbf{6.4} & 62.8 \\
        GPQA $\diamond$       & 4.9 & \textbf{6.7} & 37.7 & 5.0 & \textbf{6.4} & 27.3 \\
        LiveCodeBench      & 4.0 & \textbf{5.8} & 43.0 & 3.8 & \textbf{6.0} & 57.4 \\
        MATH-500           & 5.2 & \textbf{6.8} & 31.7 & 4.6 & \textbf{6.6} & 42.3 \\
        OlyBench-Math      & 4.5 & \textbf{6.4} & 43.9 & 4.1 & \textbf{5.9} & 43.3 \\
        OlyBench-Phy       & 3.8 & \textbf{5.6} & 48.8 & 4.2 & \textbf{5.6} & 31.6 \\
        TheoremQA          & 4.4 & \textbf{6.4} & 44.0 & 4.5 & \textbf{6.4} & 42.5 \\
        \bottomrule
        \end{tabular}
\end{table}

\subsection{Do Acceptance Length Improvements Translate To Throughput?}
\label{sec:throughput}

By default,~\ours{} updates the speculator after each speculate-verify round. While this improves acceptance length, it introduces the overhead of performing a backward pass and an optimizer step on the speculator. Thus, there exists a trade-off between acceptance length and overall throughput. We navigate this trade-off through scheduling and system-level optimizations. \textit{First}, instead of performing the gradient update in every round, we run it every $\mathcal{S}$ (update stride) rounds. Each step \textit{only} uses the latest round's training signal, without any accumulation across skipped rounds. \textit{Second}, to ensure this sporadic update does not fall on the critical path, we offload the update to a dedicated CUDA stream that runs in parallel with the next $\mathcal{S}{-}1$ speculate-verify rounds. A synchronization barrier is inserted at the end of these $\mathcal{S}{-}1$ rounds to ensure the update is complete before the speculator generates the next canvas of draft tokens. Together, these optimizations improve the applicability of \ours{} in real-world deployment settings. Please see Appendix~\ref{app:tradeoff} for more implementation details.

Table~\ref{tab:stride-sweep} shows that although $\mathcal{S}{=}1$ (lowest possible stride) achieves the highest acceptance length since the speculator is updated after each round, $\mathcal{S}{=}5$ maximizes throughput ($1.54\times$ on average and up to $1.71\times$ over DFlash) as it operates at the sweet spot in the acceptance length--throughput trade-off.

\begin{table}[!htb]
    \caption{Acceptance length and throughput of~\ours{} relative to DFlash on Qwen3.5-122B with update stride ($\mathcal{S}$). While $\mathcal{S}{=}1$ has the highest acceptance length, $\mathcal{S}{=}5$ achieves the best throughput.}
    \label{tab:stride-sweep}
    \centering
    \setlength{\tabcolsep}{5pt}
    \begin{tabular}{lcccccc}
    \toprule
    \multirow{2}{*}{Benchmark} & \multicolumn{6}{c}{Acceptance Length / Throughput (Tokens/Sec)} \\
    \cmidrule(lr){2-7}
     & DFlash & $\mathcal{S}{=}20$ & $\mathcal{S}{=}10$ & $\mathcal{S}{=}5$ & $\mathcal{S}{=}2$ & $\mathcal{S}{=}1$ \\
    \midrule
    AIME 2024       & 2.9 / 28.2 & 3.3 / 42.6 & 3.5 / 44.8 & 3.8 / \textbf{46.8} & 4.3 / 46.7 & \textbf{4.8} / 40.5 \\
    AIME 2025       & 2.9 / 27.6 & 3.7 / 43.2 & 3.8 / 44.6 & 4.3 / \textbf{45.7} & 4.7 / 45.4 & \textbf{4.9} / 37.5 \\
    GPQA $\diamond$ & 2.6 / 24.1 & 3.0 / 30.3 & 3.1 / \textbf{32.5} &  3.3 / 32.3 & 3.5 / 30.3 & \textbf{3.5} / 24.4 \\
    LiveCodeBench   & 2.4 / 24.9 & 3.1 / 31.3 & 3.2 / \textbf{31.5} & 3.4 / 31.3 & 3.8 / 30.4 & \textbf{4.0} / 25.7 \\
    MATH-500        & 3.4 / 29.7 & 3.8 / 43.7 & 4.0 / 46.5 & 4.2 / \textbf{49.7} & 4.3 / 43.5 & \textbf{4.5} / 35.9 \\
    OlyBench-Math   & 3.4 / 27.7 & 3.9 / 39.4 & 4.0 / 42.0 & 4.3 / \textbf{44.4} & 4.6 / 41.4 & \textbf{4.7} / 33.2 \\
    OlyBench-Phy    & 3.0 / 27.9 & 3.4 / 34.4 & 3.6 / \textbf{36.5} & 3.7 / 36.4 & \textbf{4.2} / 35.7 & 3.9 / 27.8 \\
    TheoremQA       & 3.3 / 30.0 & 3.7 / 41.9 & 3.7 / 42.5 & 4.1 / \textbf{51.3} & 4.1 / 44.5 & \textbf{4.2} / 30.4 \\
    \midrule
    Mean Speedup ($\times$) & 1.0 / 1.0 & 1.16 / 1.39 & 1.21 / 1.46 & 1.30 / \textbf{1.54} & 1.41 / 1.45 & \textbf{1.43} / 1.15 \\
    \bottomrule
    \end{tabular}
\end{table}

\subsection{Parameter-Efficient~\ours{} For Batched Inference Serving}
\label{sec:batching}

The experimental results so far adapt the speculator for a single in-flight request, whereas production systems serve many requests concurrently. Batched serving with~\ours{} is challenging for two reasons. \textit{First}, by default, all requests in the batch share a single set of speculator weights. As~\ours{} specializes the speculator to the current prefix per request, an update driven by one request corrupts the draft distribution of every other request in the batch. \textit{Second}, maintaining a per-request private copy of the speculator incurs significant memory overheads. Table~\ref{tab:per-memory} shows that a full-weight copy of the Qwen3.5-122B-A10B DFlash speculator along with its optimizer state costs ${\sim}2445$MB per request on each GPU at a tensor parallelism ($TP$) of four. At a batch size ($BS$) of sixty-four, this overhead scales to ${\sim}156$GB per GPU, which is $2.5{\times}$ the footprint of the \texttt{bf16} target model itself. 

\textbf{Per-request low-rank (LoRA) adapters:} We address the above challenges by freezing the speculator's pretrained weights and instead specializing through per-request LoRA adapters~(\cite{hu2021lora}). In this design,~\ours{} augments each adapted projection in the speculator with a rank-$r$ adapter and \textit{only trains the adapter's parameters}, leaving the speculator's weights \textit{untouched} and therefore \textit{safely shared across the batch}. Table~\ref{tab:per-memory} shows that specializing through LoRA adapters incurs ${\sim}86{\times}$ lower memory overhead than adapting the entire speculator. At $r{=}16$, the adaptation state per request is only $28.6$MB per GPU at the same tensor parallelism. Consequently, a batch size of sixty-four only requires $1.83$GB per GPU, which is $0.03{\times}$ the target model's footprint.

\begin{table}[H]
    \caption{Per-request adaptation state for the Qwen3.5-122B-A10B DFlash speculator at TP=4. LoRA adapters reduce adaptation state by ${\sim}86{\times}$, enabling batched serving with~\ours{}.}
    \label{tab:per-memory}
    \centering
    \begin{tabular}{lccccc}
    \toprule
    \multirow{2}{*}[-0.5\baselineskip]{Adapted Parameters} & 
    \multirowcell{2}[-2pt]{Trainable Parameters\\Per Request} &
    \multicolumn{4}{c}{Adaptation State Per GPU (GB)} \\
    \cmidrule(lr){3-6}
    & & $BS{=}8$ & $BS{=}16$ & $BS{=}32$ & $BS{=}64$ \\
    \midrule
    Full speculator weights  & $174.6$M & $19.6$ & $39.1$ & $78.2$ & $156.5$ \\
    LoRA adapters ($r{=}16$) & $2.0$M & $0.23$ & $0.46$ & $0.91$ & $1.83$ \\
    \bottomrule
    \end{tabular}
\end{table}

We also note that state-of-the-art inference engines such as SGLang~(\cite{zheng2024sglang}) and vLLM~(\cite{kwon2023efficient}) already support a similar execution model for serving distinct LoRA adapters concurrently over a shared base model~(\cite{chen2024punica, sheng2024slora}). Table~\ref{tab:batched-throughput} shows that~\ours{} with per-request LoRA adapters consistently outperforms the baseline DFlash speculator across batch sizes. We report additional batched serving results in Appendix~\ref{app:lora}.

\begin{table}[!htb]
\caption{Throughput (Tokens/Sec) of~\ours{} relative to DFlash (z-lab/Qwen3.5-122B-A10B-DFlash) on Qwen3.5-122B-A10B with $\mathcal{S}{=}10$.~Values in parentheses denote speedup over DFlash.~\ours{} consistently improves throughput by $1.5{-}2.2{\times}$ across batch sizes and benchmarks.}
\label{tab:batched-throughput}
\centering
\setlength{\tabcolsep}{4pt}
\begin{tabular}{@{}ll r@{\,}l r@{\,}l r@{\,}l r@{\,}l@{}}
    \toprule
    Benchmark & Approach & \multicolumn{2}{c}{$BS{=}8$} & \multicolumn{2}{c}{$BS{=}16$} & \multicolumn{2}{c}{$BS{=}32$} & \multicolumn{2}{c}{$BS{=}64$} \\
    \midrule
    \multirow{2}{*}{AIME 2024} 
    & DFlash & 241.2 & & 313.5 & & 423.1 & & 384.9 & \\
    & TTS~(\textit{Ours}) & \textbf{428.6} & ($1.78\times$) & \textbf{551.7} & ($1.76\times$) & \textbf{930.9} & ($2.20\times$) & \textbf{585.1} & ($1.52\times$) \\
    \midrule
    \multirow{2}{*}{AIME 2025}
    & DFlash & 213.2 & & 292.7 & & 434.8 & & 272.8 & \\
    & TTS~(\textit{Ours}) & \textbf{455.5} & ($2.14\times$) & \textbf{579.7} & ($1.98\times$) & \textbf{895.2} & ($2.06\times$) & \textbf{549.8} & ($2.02\times$) \\
    \midrule
    \multirow{2}{*}{GPQA $\diamond$}
    & DFlash & 192.0 & & 257.1 & & 428.2 & & 289.7 & \\
    & TTS~(\textit{Ours}) & \textbf{381.8} & ($1.99\times$) & \textbf{543.9} & ($2.12\times$) & \textbf{728.7} & ($1.70\times$) & \textbf{521.6} & ($1.80\times$) \\
    \midrule
    \multirow{2}{*}{LiveCodeBench}
    & DFlash & 199.3 & & 264.6 & & 378.3 & & 202.9 & \\
    & TTS~(\textit{Ours}) & \textbf{377.0} & ($1.89\times$) & \textbf{500.4} & ($1.89\times$) & \textbf{809.2} & ($2.14\times$) & \textbf{445.9} & ($2.20\times$) \\
    \bottomrule
\end{tabular}
\end{table}

\subsection{Why~\ours{} Works: A Distribution-Mismatch Perspective}

\begin{wrapfigure}{r}{0.5\textwidth}
    \vspace{-1em}
    \centering
    \includegraphics[width=0.48\textwidth]{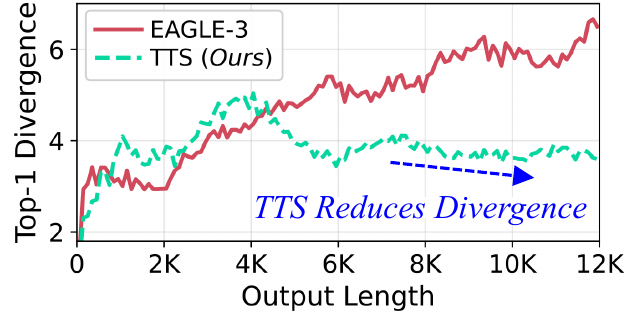}
    \caption{Cross-entropy of EAGLE-3 on Llama3.1-8B target's $Top{-}1$ token. At longer output lengths, EAGLE-3 diverges from the target;~\ours{} reduces this divergence by adapting the draft in real time.}
    \vspace{-1.5em}
    \label{fig:restore}
\end{wrapfigure}

The drafter $q(\cdot)$ is trained to approximate the target $p(\cdot)$, such that $q(\cdot \mid x) \approx p(\cdot \mid x)$ for prefixes $x$ drawn from a fixed training distribution. At inference time, however, the prefix $x_{1:t}$ grows well beyond the training distribution for real-world long-response tasks. Consequently, the approximation $q(\cdot \mid x_{1:t}) \approx p(\cdot \mid x_{1:t})$ degrades, causing a divergence between the draft and the target, as shown in Figure~\ref{fig:restore}. Each update in~\ours{} distills the current $p(\cdot \mid x_{1:t})$ back into the drafter, restoring $q_t(\cdot \mid x_{1:t}) \approx p(\cdot \mid x_{1:t})$ at the current prefix. This progressive restoration explains the compounding improvements with~\ours{} at longer generation lengths.

%% file: Sections/5_Conclusion.tex
\section{Related Work}

Prior online methods, OSD~(\cite{liu2023online}) and ATLAS~(\cite{atlas}), fine-tune the speculator using a buffer of sequences accumulated across requests and users, assuming the request stream decomposes into separable domains. In practice, however, the request stream frequently crosses domains~(\cite{zheng2023lmsys}), and a speculator tuned to past traffic falls out of alignment at every crossing, dropping acceptance length until it is re-tuned. A separate line of work retrains the speculator offline for long contexts~(\cite{yang2025longspec, lee2025owl}), but the speculator remains frozen during deployment and matches the target only on average over the training corpus. In contrast,~\ours{} adapts the draft within a single generation, resets it at the request boundary, and makes no assumption about the request stream. We present detailed comparisons in Appendix~\ref{app:related-work}.

\section{Conclusion}

State-of-the-art speculators suffer from acceptance length degradation on real-world long-response tasks, with acceptance lengths decreasing to values too close to one within a few thousand tokens and yielding negligible speedup. This arises from a gap between the speculator's training and inference distributions. We propose \textit{\underline{T}est-\underline{T}ime \underline{S}peculation (\ours{})}, an online distillation approach that closes this gap by leveraging the token verification step itself as the supervision signal needed to adapt the draft at inference time, with the target model acting as the teacher, the draft as the student, and each verification round as a distillation sample. Our evaluations across ten models and three state-of-the-art speculators show that~\ours{} improves acceptance lengths by up to $72\%$ and $41\%$ on average, with the benefits scaling with generation lengths.

%% file: Sections/6_Acknowledgments.tex
\section{Acknowledgements}

This research was conducted using computing resources on the Vista GPU Cluster through the Center for Generative AI (CGAI) and the Texas Advanced Computing Center (TACC) at the University of Texas at Austin. We thank the generous support from the Cockrell School of Engineering and the Amazon AI PhD Fellowship Program through the Amazon Science Hub at UT Austin, including the AWS cloud credits that enabled the large-scale experiments conducted in this work. Poulami Das acknowledges the generous support through the AMD endowment at UT Austin.

%% file: Sections/7_Appendix.tex
\newpage
\appendix
\titlespacing*{\section}{0pt}{3.5ex plus 1ex minus .2ex}{2.3ex plus .2ex}
\titlespacing*{\subsection}{0pt}{3.25ex plus 1ex minus .2ex}{1.5ex plus .2ex}
\titlespacing*{\subsubsection}{0pt}{3.25ex plus 1ex minus .2ex}{1.5ex plus .2ex}
\etocdepthtag.toc{appendix}
\etocsettagdepth{mainpaper}{none}
\etocsettagdepth{appendix}{subsection}
\etocsetstyle{section}{}{}{\noindent\parbox{\linewidth}{\hangindent=2.2em\hangafter=1\bfseries\etocnumber\hspace{0.9em}\etocname\mdseries\leaders\hbox to 0.6em{\hss.\hss}\hfill\etocpage}\par\vspace{9pt}}{}
\etocsetstyle{subsection}{}{}{\noindent\parbox{\linewidth}{\hangindent=4.4em\hangafter=1\hspace{2.2em}\etocnumber\hspace{0.9em}\etocname\leaders\hbox to 0.6em{\hss.\hss}\hfill\etocpage}\par\vspace{6pt}}{}
\etocsettocstyle{}{}
\onecolumn
\thispagestyle{empty}
\begin{center}
{\fontsize{20pt}{24pt}\selectfont\bfseries Appendix\par}
\vspace{6pt}
{\fontsize{11pt}{13pt}\selectfont\itshape Table of Contents\par}
\end{center}
\vspace{18pt}
{\fontsize{11pt}{14pt}\selectfont
\tableofcontents
}

\newpage
\section{\ours{} Generalizes Across Speculators}
\label{app:spec-results}

\ours{} generalizes across speculators. To demonstrate this, we evaluate the performance of~\ours{} across several models for both EAGLE-3 and PARD. Table~\ref{tab:eagle-comp} shows that~\ours{} consistently outperforms vanilla EAGLE-3 across all benchmarks and models, improving acceptance length by up to $66.8\%$.

\begin{table}[H]
    \caption{Acceptance length of EAGLE-3 (EGL-3) and~\ours{} across models and tasks. $\Delta(\%)$ denotes the relative improvement of~\ours{} over EGL-3.~\ours{} consistently achieves higher acceptance lengths.}
    \label{tab:eagle-comp}
    \centering
    \begin{tabular}{l @{\hskip 0em} c@{\hskip 3pt}c@{\hskip 3pt}c @{\hskip 0.5em} c@{\hskip 3pt}c@{\hskip 3pt}c @{\hskip 0.5em} c@{\hskip 3pt}c@{\hskip 3pt}c}
    \toprule
    \multirow{2}{*}{Dataset} 
    & \multicolumn{3}{c}{Llama3.1-8B} 
    & \multicolumn{3}{c}{Qwen/Qwen3-8B} 
    & \multicolumn{3}{c}{Qwen/Qwen3-32B} \\
    \cmidrule(lr){2-4} \cmidrule(lr){5-7} \cmidrule(lr){8-10}
    & EGL-3 & TTS(\textit{Ours}) & $\Delta(\%)$ & EGL-3 & TTS(\textit{Ours}) & $\Delta(\%)$ & EGL-3 & TTS(\textit{Ours}) & $\Delta(\%)$ \\
    \midrule
    AIME 2024        & 1.2 & \textbf{2.0} & 64.7 & 1.6 & \textbf{1.9} & 18.7 & 1.4 & \textbf{1.9} & 26.3 \\
    AIME 2025        & 1.2 & \textbf{1.9} & 63.7 & 1.6 & \textbf{1.9} & 20.4 & 1.5 & \textbf{1.9} & 25.5 \\
    GPQA $\diamond$  & 1.2 & \textbf{2.0} & 60.4 & 1.5 & \textbf{1.9} & 24.5 & 1.4 & \textbf{1.8} & 21.5 \\
    LiveCodeBench    & 1.4 & \textbf{2.1} & 47.0 & 1.7 & \textbf{1.9} & 12.7 & 1.5 & \textbf{1.8} & 22.1 \\
    MATH-500         & 1.2 & \textbf{2.0} & 61.8 & 1.6 & \textbf{1.9} & 18.8 & 1.5 & \textbf{1.8} & 21.5 \\
    OlyBench-Math    & 1.3 & \textbf{2.1} & 66.8 & 1.6 & \textbf{1.9} & 18.5 & 1.5 & \textbf{1.9} & 24.5 \\
    OlyBench-Phy     & 1.2 & \textbf{1.9} & 65.4 & 1.6 & \textbf{1.9} & 17.2 & 1.5 & \textbf{1.8} & 21.6 \\
    TheoremQA        & 1.2 & \textbf{1.9} & 56.9 & 1.6 & \textbf{1.9} & 16.8 & 1.5 & \textbf{1.8} & 21.9 \\
    \bottomrule
    \end{tabular}
\end{table}

Table~\ref{tab:pard-comp} shows the performance of~\ours{} against PARD. We observe that~\ours{} consistently outperforms vanilla PARD across all benchmarks and models, improving acceptance length by up to $272.2\%$.

\begin{table}[H]
    \caption{Acceptance length of~\ours{} and PARD across models and tasks. $\Delta(\%)$ denotes the relative improvement of~\ours{} over PARD.~\ours{} consistently achieves higher acceptance lengths.}
    \label{tab:pard-comp}
    \centering
    \begin{tabular}{l @{\hskip 0em} c@{\hskip 3pt}c@{\hskip 3pt}c @{\hskip 0.5em} c@{\hskip 3pt}c@{\hskip 3pt}c @{\hskip 0.5em} c@{\hskip 3pt}c@{\hskip 3pt}c}
    \toprule
    \multirow{2}{*}{Dataset}
    & \multicolumn{3}{c}{DeepSeek-R1-Qwen-7B}
    & \multicolumn{3}{c}{DeepSeek-R1-Qwen-14B}
    & \multicolumn{3}{c}{DeepSeek-R1-Qwen-32B} \\
    \cmidrule(lr){2-4} \cmidrule(lr){5-7} \cmidrule(lr){8-10}
    & PARD & TTS(\textit{Ours}) & $\Delta(\%)$ & PARD & TTS(\textit{Ours}) & $\Delta(\%)$ & PARD & TTS(\textit{Ours}) & $\Delta(\%)$ \\
    \midrule
    AIME 2024       & 3.0 & {\bf 6.0} & 100.0 & 3.2 & {\bf 5.3} & 65.6  & 3.4 & {\bf 5.5} & 61.8  \\
    AIME 2025       & 2.9 & {\bf 5.9} & 103.4 & 2.7 & {\bf 4.6} & 70.4  & 2.9 & {\bf 4.8} & 65.5  \\
    GPQA $\diamond$ & 2.4 & {\bf 5.6} & 133.3 & 3.2 & {\bf 5.4} & 68.8  & 2.8 & {\bf 6.1} & 117.9 \\
    LiveCodeBench   & 1.8 & {\bf 6.7} & 272.2 & 2.0 & {\bf 6.1} & 205.0 & 1.9 & {\bf 6.1} & 221.1 \\
    MATH-500        & 4.2 & {\bf 5.7} & 35.7  & 3.9 & {\bf 5.0} & 28.2  & 4.0 & {\bf 5.1} & 27.5  \\
    OlyBench-Math   & 2.0 & {\bf 6.0} & 200.0 & 2.6 & {\bf 5.2} & 100.0 & 2.3 & {\bf 5.5} & 139.1 \\
    OlyBench-Phy    & 2.0 & {\bf 6.1} & 205.0 & 2.2 & {\bf 5.2} & 136.4 & 2.3 & {\bf 5.7} & 147.8 \\
    TheoremQA       & 3.8 & {\bf 5.4} & 42.1  & 3.9 & {\bf 5.0} & 28.2  & 4.0 & {\bf 5.1} & 27.5  \\
    \bottomrule
    \end{tabular}
\end{table}

\newpage
\section{Benchmarks Used For Evaluation}
\label{app:benchmarks}

All benchmarks used to evaluate~\ours{} elicit long outputs, often generating tens of thousands of tokens before reaching the final answer, either within a single response or across a multi-turn session. These benchmarks collectively represent long-generation tasks that dominate real-world workloads. For each benchmark, we randomly select 32 sequences (except AIME, which comprises only 30 sequences) using a fixed seed for deterministic comparison. We consolidate the identifiers for each benchmark in Table~\ref{tab:benchmarks} and describe them in detail below:

\begin{table}[htp]
\centering
\caption{HuggingFace identifiers for benchmarks used in our evaluations, grouped by domain.}
\label{tab:benchmarks}
\begin{tabular}{l l}
\toprule
\textbf{Benchmark} & \textbf{HuggingFace Dataset Identifier} \\
\midrule
\multicolumn{2}{l}{\em Competition mathematics} \\
AIME 2024            & \texttt{MathArena/aime\_2024\_I, MathArena/aime\_2024\_II} \\
AIME 2025            & \texttt{MathArena/aime\_2025} \\
MATH-500             & \texttt{HuggingFaceH4/MATH-500 (Level 5)} \\
OlympiadBench-Math   & \texttt{Hothan/OlympiadBench} (\texttt{OE\_TO\_maths\_en\_COMP}) \\
\midrule
\multicolumn{2}{l}{\em Physics} \\
OlympiadBench-Physics & \texttt{Hothan/OlympiadBench} (\texttt{OE\_TO\_physics\_en\_COMP}) \\
\midrule
\multicolumn{2}{l}{\em Graduate-level scientific reasoning} \\
GPQA $\diamond$      & \texttt{Idavidrein/gpqa} (\texttt{gpqa\_diamond}) \\
\midrule
\multicolumn{2}{l}{\em University-level theorem application} \\
TheoremQA            & \texttt{TIGER-Lab/TheoremQA} \\
\midrule
\multicolumn{2}{l}{\em Code generation} \\
LiveCodeBench v6     & \texttt{livecodebench/code\_generation\_lite (Hard)} \\
\midrule
\multicolumn{2}{l}{\em Multi-turn agentic tasks} \\
OpenThoughts-Agent   & \texttt{open-thoughts/OpenThoughts-Agent-SFT-100K} \\
\bottomrule
\end{tabular}
\end{table}

\textbf{AIME:} The AIME dataset collects the American Invitational Mathematics Examination problems from each year. Both the AIME 2024~(\cite{aime2024_I,aime2024_II}) and the AIME 2025~(\cite{aime2025}) comprise 30 sequences with a final integer answer in $[0, 999]$. The AIME dataset is a strong test for olympiad-style reasoning as it elicits long outputs of up to $32K$ tokens.

\textbf{MATH-500:} The MATH-500 dataset is a curated $500$-problem subset of the MATH benchmark~(\cite{hendrycks2021measuring}), introduced in OpenAI's PRM800K~(\cite{lightman2023let}) for evaluating mathematical-reasoning models. Problems are drawn from several competitions such as AMC, and span seven subject areas with difficulty levels $1$ through $5$. We use only the Level 5 problems, the most challenging subset.

\textbf{OlympiadBench-Math:} OlympiadBench~(\cite{he2024olympiadbench}) is a bilingual olympiad-style benchmark drawn from international and Chinese olympiad competitions. We use English-language problems from this dataset in our evaluations. Furthermore, we restrict evaluation to text-only sequences as not all evaluation models are multi-modal. Each sequence is accompanied by a symbolic or numeric final answer. Similar to AIME and MATH-500, OlympiadBench-Math elicits long generation lengths.

\textbf{OlympiadBench-Physics:} OlympiadBench-Physics~(\cite{he2024olympiadbench}) is the physics counterpart of OlympiadBench-Math, drawn from the same olympiad-competition pipeline and similarly restricted to its English-language subset. Each problem requires deriving a closed-form or numeric answer from first principles, often achieved via multi-step modeling and symbolic manipulation.

\textbf{GPQA $\diamond$:} GPQA $\diamond$~(\cite{rein2023gpqa}) is the hardest $198$-question subset of the GPQA benchmark. It comprises graduate-level multiple-choice questions in biology, chemistry, and physics. PhD-level domain experts answer GPQA $\diamond$ correctly only ${\sim}65\%$ of the time, while non-experts achieve ${\leq}35\%$ accuracy even with unrestricted web access, making it a strong test of expert-level scientific reasoning. 

\textbf{TheoremQA:} TheoremQA~(\cite{chen2023theoremqa}) is a STEM benchmark consisting of sequences that require identifying and applying a specific named theorem from mathematics, physics, or engineering to produce a numeric or symbolic final answer. TheoremQA is a strong test of theorem application, and elicits long reasoning traces as the model recalls, applies, and verifies the relevant result.

\textbf{LiveCodeBench:} LiveCodeBench~(\cite{jain2024livecodebench}) is a competitive-programming benchmark drawn from LeetCode, AtCoder, and CodeForces. Each sequence includes the problem statement, the starter code, and an explicit instruction to reason step by step before emitting a solution.

\textbf{OpenThoughts-Agent:} OpenThoughts-Agent-SFT-100K~(\cite{openthoughts-agent}) is a collection of $100K$ agentic tasks paired with reference trajectories, sourced from SWE-Smith, StackExchange, and GitHub issue resolution. We use only the task specifications and discard the reference trajectories, since~\ours{} adapts the speculator to the trajectory produced by the target model itself. Each task is executed in a sandboxed shell environment under the Terminus-2 agent protocol~(\cite{merrill2026terminal}), where the model interleaves tool calls with environment observations until the task is complete. Unlike the single-turn benchmarks above, generation here is distributed across an entire session spanning many turns, which collectively elicits tens of thousands of output tokens.

\section{Evaluated Models And Speculators}
\label{app:models}

We use ten state-of-the-art models across the Qwen-3~(\cite{yang2025qwen3}), Qwen-3.5~(\cite{qwen3.5}), DeepSeek-R1-Distill-Qwen~(\cite{guo2025deepseek}), GPT-OSS~(\cite{openai2025gptoss120bgptoss20bmodel}), MiniMax~(\cite{chen2026minimax}), and Llama3.1~(\cite{grattafiori2024llama}) families. Our selection of models encompasses parameter sizes from 4 billion to 229 billion. The Qwen3 models (4B, 8B, 32B), Llama3.1-8B-Instruct, and the DeepSeek-R1-Distill-Qwen models (7B, 14B, 32B) are dense transformer models. GPT-OSS-120B, Qwen3.5-122B-A10B, and MiniMax-M2.5 are mixture-of-experts models~(\cite{shazeer2017outrageously}). The first two interleave full-attention layers with cheaper alternatives, namely sliding-window attention in GPT-OSS-120B and linear attention in Qwen3.5-122B-A10B, whereas MiniMax-M2.5 uses full attention in every layer. Table~\ref{tab:models} summarizes the specifications of each model.

\begin{table}[H]
  \centering
  \setlength{\tabcolsep}{3.5pt}
  \caption{Our selection of models spans parameter sizes from 4B to 229B and encompasses dense, MoE, and hybrid MoE architectures across six model families.}
  \label{tab:models}
  \begin{tabular}{l c c c c}
  \toprule
  \textbf{HuggingFace Identifier} & \textbf{Architecture} & \makecell{\textbf{Active /}\\\textbf{Total}} & \textbf{Layers} & \makecell{\textbf{Context}\\\textbf{Length}} \\
  \midrule
  Qwen/Qwen3-4B          & Dense        & 4B / 4B     & 36 & $40$K  \\
  Qwen/Qwen3-8B          & Dense        & 8B / 8B     & 36 & $40$K  \\
  Qwen/Qwen3-32B         & Dense        & 32B / 32B   & 64 & $40$K  \\
  deepseek-ai/DeepSeek-R1-Distill-Qwen-7B  & Dense & 7B / 7B   & 28 & $128$K \\
  deepseek-ai/DeepSeek-R1-Distill-Qwen-14B & Dense & 14B / 14B & 48 & $128$K \\
  deepseek-ai/DeepSeek-R1-Distill-Qwen-32B & Dense & 32B / 32B & 64 & $128$K \\
  meta-llama/Llama-3.1-8B-Instruct         & Dense        & 8B / 8B     & 32 & $128$K \\
  openai/gpt-oss-120b & MoE & 5.1B / 117B & 36 & $128$K \\
  Qwen/Qwen3.5-122B-A10B & Hybrid + MoE & 10B / 122B  & 48 & $256$K \\
  MiniMaxAI/MiniMax-M2.5 & MoE & 10B / 229B & 62 & $192$K \\
  \bottomrule
  \end{tabular}
\end{table}

For each model in Table~\ref{tab:models}, we evaluate~\ours{} on top of the matching publicly released DFlash~(\cite{chen2026dflash}), EAGLE-3~(\cite{li2025eagle}), and PARD~(\cite{an2025pard}) checkpoints. We describe each speculator below in detail and summarize their HuggingFace identifiers in Table~\ref{tab:speculators}.
    
\textbf{DFlash:} DFlash is a \textit{block-diffusion} speculator that generates a block of $B$ tokens in a single forward pass. It first extracts the intermediate hidden states of a selected subset of target model layers and then passes them through the speculator's $W_{k}$ and $W_{v}$ projection matrices to produce the context Key-Value (KV) cache. Next, this context KV cache is concatenated with the speculator's own KV vectors for the block of tokens being generated. Therefore, the query vectors attend to the target context and the block context in a single forward pass.

\textbf{EAGLE-3:} EAGLE-3 is an \textit{auto-regressive} speculator that generates one token at a time. It first extracts the hidden states from three layers (typically, an early, a middle, and a late layer) of the target model. These hidden states are then concatenated and down-projected through a fully connected layer to produce a $k$-dimensional context feature, where $k$ represents the hidden state dimension of the target model. This context feature is then concatenated with the embedding vector of the previously generated output token, reduced through a second fully connected layer and fed into the EAGLE-3 draft, a single transformer layer. Its output is then passed to the LM head of the target model to produce the next-token distribution. 

\textbf{PARD:} PARD is a \textit{parallel} speculator that generates a block of $K$ tokens in a single forward pass. It appends $K$ mask tokens to the current sequence, one for each draft token to be generated. The speculator then predicts the token at every masked position simultaneously, where each prediction conditions on the committed prefix and the preceding mask tokens, allowing all $K$ tokens to be generated in parallel. Unlike EAGLE-3 and DFlash, the PARD speculator is a \textit{standalone} model that does not reuse the target model's hidden states. It adapts an existing auto-regressive model offline to predict the masked tokens in parallel through fine-tuning.

We use a block size of $16$ and $8$ for DFlash and PARD, respectively, and a generation depth of $8$ for EAGLE-3, across all target models, consistent with the original papers~(\cite{chen2026dflash, an2025pard, li2025eagle}). We use the publicly released speculator checkpoints from the original authors, except for EAGLE-3 on the Qwen3 and GPT-OSS target models, for which the authors have not released checkpoints; there, we use the AngelSlim and NVIDIA releases from HuggingFace, respectively. Table~\ref{tab:speculators} summarizes the speculator checkpoints used in our evaluations.

\begin{table}[H]
    \centering
    \caption{HuggingFace identifiers for each speculator and target model pair.}
    \label{tab:speculators}
    \begin{threeparttable}
    \begin{tabular}{c c c}
    \toprule
    \textbf{Target Model} & \textbf{Speculator} & \textbf{HuggingFace Identifier} \\
    \midrule
    meta-llama/Llama-3.1-8B-Instruct & \multirow{5}{*}{EAGLE-3} & yuhuili/EAGLE3-LLaMA3.1-Instruct-8B \\
    Qwen/Qwen3-8B           &    & AngelSlim/Qwen3-8B\_eagle3 \\
    Qwen/Qwen3-32B          &    & AngelSlim/Qwen3-32B\_eagle3 \\
    openai/gpt-oss-120b     &    & nvidia/gpt-oss-120b-Eagle3-v3 \\
    openai/gpt-oss-120b     &    & nvidia/gpt-oss-120b-Eagle3-long-context \\
    \midrule
    Qwen/Qwen3-4B           & \multirow{4}{*}{DFlash} & z-lab/Qwen3-4B-DFlash-b16 \\
    Qwen/Qwen3-8B           &     & z-lab/Qwen3-8B-DFlash-b16 \\
    Qwen/Qwen3.5-122B-A10B  &     & z-lab/Qwen3.5-122B-A10B-DFlash \\
    MiniMaxAI/MiniMax-M2.5  &     & z-lab/MiniMax-M2.5-DFlash \\ 
    \midrule
    DeepSeek-R1-Distill-Qwen-7B\tnote{*}  &  \multirow{3}{*}{PARD}  & \multirow{3}{*}{amd/PARD-DeepSeek-R1-Distill-Qwen-1.5B}\\
    DeepSeek-R1-Distill-Qwen-14B\tnote{*} &    & \\
    DeepSeek-R1-Distill-Qwen-32B\tnote{*} &    & \\
    \bottomrule
    \end{tabular}
    \begin{tablenotes}
    \item[*] Hosted under the \texttt{deepseek-ai/} organization on HuggingFace.
    \end{tablenotes}
    \end{threeparttable}
\end{table}

\clearpage
\section{Implementation Details Of~\ours{}}
\label{app:tradeoff}

In this section, we provide the implementation details of~\ours{}, including scheduling and system-level optimizations introduced in Section~\ref{sec:throughput}. We then describe the system-level challenges we encountered in integrating~\ours{} into the SGLang inference framework.

\subsection{Navigating The Acceptance Length--Throughput Trade-Off}
Every gradient update in~\ours{} invokes a backward pass and an optimizer (Adam) step on the speculator. While this improves acceptance length by continuously adapting the speculator to match the target distribution, it introduces additional overhead on the critical path in every speculate-verify round. The latency overhead of a single backward pass and optimizer step is comparable to a few speculation rounds. Thus, applying~\ours{} naively can negate the throughput improvements from higher acceptance length. We navigate this trade-off via scheduling and system-level optimizations as described below:

\noindent \textbf{Strided Updates:} By default,~\ours{} performs a gradient update after every speculate-verify round. A natural alternative is to perform the gradient update only once every $\mathcal{S}$ rounds, where $\mathcal{S}$ is the \textit{update stride}. Each update consumes the supervision signal only from the most recent round, without any accumulation across the $\mathcal{S}{-}1$ skipped rounds. However, fewer updates imply weaker adaptation to the ongoing generation and therefore reduce the improvements in acceptance length. For example, Table~\ref{tab:freq-benchmark} shows that the improvements in acceptance length are the highest for $\mathcal{S}{=}1$ and monotonically decrease as the update stride is increased.

\begin{table}[H]
    \caption{Acceptance length of~\ours{} relative to DFlash on Qwen3-8B with the update stride $\mathcal{S}$. The default $\mathcal{S}{=}1$ achieves the largest improvement across benchmarks. Higher values of $\mathcal{S}$ reduce the improvement in acceptance length as the speculator is adapted less frequently.}
    \label{tab:freq-benchmark}
    \centering
    \begin{tabular}{lcccccc}
    \toprule
    \multirow{2}{*}{Benchmark} & \smash{\raisebox{-14pt}{\shortstack{DFlash\\[0pt](\textit{Baseline})}}} & \multicolumn{5}{c}{TTS (\textit{Ours}) / $\Delta(\%)$} \\
    \cmidrule(lr){3-7}
    &  & $\mathcal{S}{=}1$\textit{(default)} & $\mathcal{S}{=}5$ & $\mathcal{S}{=}15$ & $\mathcal{S}{=}30$ & $\mathcal{S}{=}50$ \\
    \midrule
    AIME 2024       & 3.3 & \textbf{4.6} / 38.5 & 4.3 / 30.7 & 3.9 / 19.0 & 3.7 / 12.6 & 3.6 / 10.2 \\
    AIME 2025       & 3.1 & \textbf{4.1} / 33.1 & 3.9 / 27.0 & 3.7 / 18.1 & 3.5 / 12.1 & 3.3 /  8.0 \\
    GPQA $\diamond$ & 3.2 & \textbf{4.7} / 47.3 & 4.3 / 34.3 & 4.0 / 24.1 & 3.8 / 17.8 & 3.7 / 14.7 \\
    LiveCodeBench   & 2.9 & \textbf{4.5} / 55.2 & 4.2 / 44.9 & 3.9 / 32.9 & 3.6 / 24.4 & 3.5 / 19.8 \\
    MATH-500        & 3.8 & \textbf{4.3} / 13.3 & 4.1 /  8.3 & 4.0 /  5.4 & 3.9 /  3.7 & 3.9 /  2.3 \\
    OlyBench-Math   & 2.7 & \textbf{4.1} / 50.3 & 3.8 / 42.1 & 3.5 / 28.9 & 3.2 / 20.0 & 3.1 / 14.2 \\
    OlyBench-Phy    & 2.8 & \textbf{4.3} / 53.1 & 3.9 / 39.2 & 3.5 / 26.2 & 3.3 / 18.2 & 3.2 / 13.3 \\
    TheoremQA       & 3.5 & \textbf{4.3} / 22.8 & 4.0 / 15.2 & 3.8 /  9.0 & 3.7 /  6.3 & 3.7 /  4.8 \\
    \bottomrule
    \end{tabular}
\end{table}

However, increasing the stride $\mathcal{S}$ amortizes the update overhead, since the gradient step is now performed every $\mathcal{S}$ rounds rather than at every round. For example, Table~\ref{tab:qwen3-livecodebench} shows that the best wall-clock speedup ($\mathbf{1.83\times}$) is achieved at $\mathcal{S}{=}10$. Beyond $\mathcal{S}{=}10$, speedup decreases because the degradation of acceptance length outpaces the savings from amortizing the gradient update.

\begin{table}[H]
    \caption{Relative acceptance length (AL) and wall-clock speedup of~\ours{} on LiveCodeBench with Qwen3-8B. Both metrics are reported relative to $\mathcal{S}{=}1$. The sweet spot is at $\mathcal{S}{=}10$ ($\mathbf{1.83\times}$ speedup) where the update overhead is amortized while acceptance length is preserved within $0.88\times$ of $\mathcal{S}{=}1$.}
    \label{tab:qwen3-livecodebench}
    \centering
    \begin{tabular}{lcccccccc}
    \toprule
    \multirow{2}{*}{Metric} & \multicolumn{8}{c}{Update Stride ($\mathcal{S}$)} \\
    \cmidrule(lr){2-9}
    & 1\textit{(default)} & 5 & 10 & 15 & 20 & 30 & 40 & 50 \\
    \midrule
    Relative AL & 1.00$\times$ & 0.93$\times$ & 0.88$\times$ & 0.85$\times$ & 0.84$\times$ & 0.80$\times$ & 0.79$\times$ & 0.77$\times$ \\
    Relative Speedup & 1.00$\times$ & 1.77$\times$ & \textbf{1.83$\times$} & 1.79$\times$ & 1.80$\times$ & 1.65$\times$ & 1.64$\times$ & 1.56$\times$ \\
    \bottomrule
    \end{tabular}
\end{table}

Although strided updates reduce the frequency of the gradient steps, each update still lies on the critical path of the round where it is performed. For small target models, the cost of one gradient update to the speculator is comparable to a few speculation rounds and thus can degrade wall-clock speedup. We further hide this latency through \textit{asynchronous pipelining}, described next.

\noindent \textbf{Asynchronous Pipelining:} If the gradient update could be performed in parallel with the subsequent speculation rounds, we could hide its latency (largely) from the critical path. Building on this insight, we design mechanisms in~\ours{} to offload the gradient update to a dedicated CUDA stream (the \textit{side stream}) that executes concurrently with the next $\mathcal{S}{-}1$ speculation rounds on the main stream. However, asynchronous pipelining does not fully eliminate the latency overhead of the gradient update. This is because the side stream and the main stream contend for the same GPU resources, slightly slowing the speculation rounds that run concurrently with a gradient update. Nonetheless, the additional overhead is much smaller than executing the gradient update on the critical path. A synchronization barrier is inserted after $\mathcal{S}{-}1$ rounds to ensure that the speculator is adjusted with the prior update before the next canvas of draft tokens is generated.

\begin{figure*}[htp]
\begin{center}
    \centering
    \vspace{-0.05in}
    \includegraphics[width=1.0\textwidth]{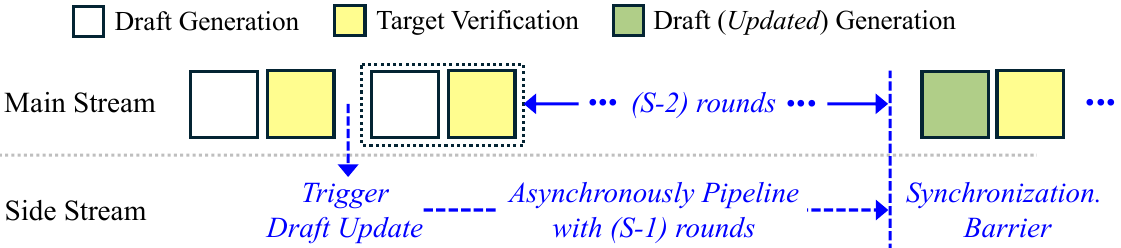}
    \caption{Execution timeline of~\ours{} with strided updates and asynchronous pipelining. Every $\mathcal{S}$ speculation rounds, a gradient update is dispatched to a side CUDA stream and overlaps with the next $\mathcal{S}{-}1$ rounds on the main stream, keeping the update off the critical path.}
    \vspace{-0.05in}
\label{fig:implementation}
\end{center}
\end{figure*}

Figure~\ref{fig:implementation} illustrates the resulting execution timeline of~\ours{} with both these optimizations applied. 
 
\subsection{Limitations And Future Work}
While~\ours{} delivers substantial speedup, the achievable throughput remains limited by two missing system-level optimizations. \textit{First}, the gradient update to adapt the speculator is currently implemented as a sequence of standard PyTorch kernels. Fusing these into a custom kernel would reduce the latency of each update and improve throughput across all strides. \textit{Second}, the update stride is currently fixed and applied statically across all speculation rounds. We could, however, adapt the stride online based on the observed acceptance length, automatically choosing a lower stride when the acceptance length degrades and a higher stride when it remains stable. Both these optimizations are orthogonal to the main contributions of this paper, and therefore we defer them to future work.

\subsection{System-Level Challenges In SGLang Integration}
SGLang is a framework designed for fast inference rather than online adaptation. We describe below the two primary challenges of integrating~\ours{} into SGLang, along with their resolutions:

\begin{enumerate}
    \item \textbf{Inference Kernels Do Not Support Gradient Computation:} To maximize throughput, SGLang uses fused CUDA/Triton kernels for attention, normalization, and activation. These kernels are not designed for gradient computation. During the backward pass,~\texttt{autograd} breaks silently. As a result, the gradient update has no effect on the speculator weights. We resolve this by switching to the differentiable PyTorch implementation for the speculator in the update round, ensuring that the gradients flow correctly through the backward pass. On non-update rounds, we retain the fused kernels to preserve throughput.
    \item \textbf{CUDA-Graph Replay Does Not Honor Cross-Stream Synchronization:} To improve performance, each draft generation round in SGLang runs inside a captured CUDA graph, a pre-recorded sequence of GPU operations replayed for efficiency. Graph replay runs on its own internal stream and does not honor GPU synchronization barriers. Thus, the update made by~\ours{} can be silently skipped before the next round begins. We resolve this by introducing a CPU-side synchronization barrier that blocks the GPU until the side stream completes the update before the next graph is replayed. This barrier is inserted only at update boundaries, leaving the CUDA graph performance unaffected on non-update rounds.
\end{enumerate}

These optimizations and resolutions enable~\ours{} to run end-to-end on SGLang.

\clearpage
\section{Related Work}
\label{app:related-work}

\subsection{Online Adaptation Methods}
\label{app:related-online}

In this section, we compare and contrast~\ours{} with prior online methods that adapt the speculator after deployment. These works adapt the speculator across requests and users, and assume that production traffic decomposes into cleanly separable domains.

Online speculative decoding (OSD)~(\cite{liu2023online}) periodically fine-tunes the draft model to match the target distribution on a buffer of accumulated sequences \textit{across users and requests}. The draft is updated when the buffer exceeds a certain size. OSD additionally orchestrates multiple speculators, each specialized to a particular domain, and uses a router to direct each incoming request to the best-matching speculator. Each draft is only fine-tuned on the requests it receives. However, this multi-draft \textit{setup diverges} from how state-of-the-art speculators are typically deployed in practice~(\cite{tang2025efficient}), where a single speculator (such as EAGLE-3 or DFlash) is paired with the target model and serves all incoming queries.

ATLAS~(\cite{atlas}) extends this paradigm to a production serving system. Rather than routing between domain-specialized speculators, ATLAS pairs a heavyweight static speculator (Together AI's proprietary Turbo speculator) with a lightweight adaptive speculator that is updated on production traffic. A confidence-aware controller chooses which to trust at each step. However, the lightweight speculator is still adapted across requests over accumulated sequences. Consequently, its adaptation reflects the aggregate of recent traffic rather than the request currently being served.

This cross-request adaptation, however, relies on the \textit{assumption that production traffic cleanly decomposes into separable domains}. The evaluation of OSD structures the request stream as a concatenation of single-domain benchmarks. Specifically, OSD evaluates on streams of $2K$ sequences each from Spider (SQL)~(\cite{yu2018spider}), GSM8K (math)~(\cite{gsm8k}), Code-search-Python (code)~(\cite{husain2019codesearchnet}), and Alpaca-finance (finance)~(\cite{bharti2023alpacafinance}), concatenated sequentially before the stream switches to the next benchmark. Similarly, ATLAS adapts its lightweight speculator on accumulated live traffic, aligning it to the aggregate distribution of recent requests rather than the current request being served. In practice, however, request streams cross domain boundaries frequently~(\cite{zheng2023lmsys, sharegpt}), and a speculator tuned to the aggregate of recent traffic falls out of alignment at every crossing. Furthermore, \textit{neither method demonstrates compatibility} with public state-of-the-art speculators such as EAGLE-3, DFlash, and PARD; OSD relies on an auxiliary draft model, while ATLAS utilizes Together AI's proprietary Turbo speculator.

Fundamentally, prior works adapt the speculator at the granularity of traffic accumulated across many requests, whereas acceptance length is determined by how closely the draft distribution ($q(\cdot \mid x_{1:t})$) matches the target ($p(\cdot \mid x_{1:t})$) for the current prefix ($x_{1:t}$). Adapting to the aggregate of recent traffic improves the speculator only when the current request resembles the past request stream, and cannot specialize to the request being served. In contrast,~\ours{} adapts the speculator within each generation at the granularity of each speculation round (or every $\mathcal{S}$ rounds with strided updates). It distills the target's per-prefix conditional $p(\cdot \mid x_{1:t})$ back into the speculator at every update and resets it between requests, so per-request adaptation is bounded. It also operates directly on top of public state-of-the-art speculators. Moreover,~\ours{} makes no assumption about the structure of the request stream and consistently improves acceptance length across benchmarks. Furthermore, to improve applicability, we implement~\ours{} in the production-grade SGLang inference serving engine~(\cite{zheng2024sglang}). Table~\ref{tab:related-comparison} summarizes the key differences between~\ours{} and prior work.

\begin{table}[H]
\centering
\caption{Comparison of online speculator adaptation methods.}
\label{tab:related-comparison}
\setlength{\tabcolsep}{6pt}
\begin{tabular}{l c c c}
\toprule
\textbf{Comparison Property} & \textbf{OSD} & \textbf{ATLAS} & \textbf{\ours{}} \\
\midrule
Adapts within a single generation & \xmark & \xmark & \cmark \\
No assumption on the request stream structure & \xmark & \xmark & \cmark \\
Prevents cross-request leakage & \xmark & \xmark & \cmark \\
Compatible with state-of-the-art speculators (EAGLE-3, DFlash) & \xmark & \xmark & \cmark \\
Single-speculator deployment (no specialized routing) & \xmark & \cmark & \cmark \\
No proprietary dependencies & \cmark & \xmark & \cmark \\
Production-grade serving integration & \xmark & \cmark & \cmark \\
\bottomrule
\end{tabular}
\end{table}

\subsubsection{Quantitative Comparison Against Online Adaptation Methods}
\label{app:related-online-quant}

To demonstrate the efficacy of the per-request adaptation in~\ours{} against the cross-request adaptation used by prior works, we implement and integrate OSD in SGLang and compare its performance against~\ours{} on the DFlash speculator. 

Following the official public codebase\footnote{\url{https://github.com/LiuXiaoxuanPKU/OSD}}, we perform one gradient update on the speculator at an interval of eight requests ($I{=}8$), minimizing the soft cross-entropy between the speculator's and target's next-token distributions at the positions where the target rejected the speculator's proposed tokens. For this comparison, both OSD and~\ours{} train the entire speculator and use an identical learning rate of \texttt{1e-4}. Thus, both methods only differ in the granularity and persistence of adaptation. We evaluate both approaches on the Qwen3.5-122B-A10B target and its DFlash speculator (\texttt{z-lab/Qwen3.5-122B-A10B-DFlash}), serving an identical request stream of four benchmarks: LiveCodeBench, MATH-500, OlympiadBench-Math (zh), and OlympiadBench-Physics (zh), with requests issued in short bursts of two per domain. 

Table~\ref{tab:osd-comp} shows that~\ours{} achieves higher acceptance length than OSD across all benchmarks. As OSD \textit{persists} the speculator state across requests, the measured acceptance length is strongly influenced by the preceding requests. Furthermore, as adaptation in OSD is \textit{unbounded}, the speculator's weights drift further from their pretrained values with each gradient update. In contrast,~\ours{} adapts within each request and resets the speculator at the request boundary, so its adaptation always reflects the current request and cannot degrade any subsequent one.

\begin{table}[H]
    \caption{Mean acceptance length (AL) on an interleaved request stream for the Qwen3.5-122B-A10B model and its DFlash speculator. While OSD adapts across requests and cannot specialize to the request being served,~\ours{} resets at the request boundary and consistently achieves higher AL.}
    \label{tab:osd-comp}
    \centering
    \begin{tabular}{lccc}
    \toprule
    \multirow{2}{*}{Benchmark} & \multicolumn{3}{c}{Acceptance Length} \\
    \cmidrule(lr){2-4}
    & DFlash & OSD & TTS~(\textit{Ours}) \\
    \midrule
    LiveCodeBench              & 2.06 & 2.50 ($+$21.4\%) & \textbf{3.57} ($+$73.3\%) \\
    MATH-500                   & 4.66 & 5.38 ($+$15.5\%) & \textbf{5.49} ($+$17.8\%) \\
    OlyBench-Math (zh)    & 3.11 & 3.84 ($+$23.5\%) & \textbf{4.72} ($+$51.8\%) \\
    OlyBench-Physics (zh) & 3.30 & 3.78 ($+$14.5\%) & \textbf{4.25} ($+$28.8\%) \\
    \midrule
    Mean                       & 3.28 & 3.88 ($+$18.1\%) & \textbf{4.51} ($+$37.3\%) \\
    \bottomrule
    \end{tabular}
\end{table}

\subsection{Offline Adaptation Methods}
\label{app:related-offline}

In this section, we compare and contrast~\ours{} with prior offline methods that improve long-context speculation before deployment. These works redesign and retrain the speculator offline on long-context data to reduce the training-inference mismatch described in Section~\ref{sec:diagnosis}.

LongSpec~(\cite{yang2025longspec}) addresses the training-inference data mismatch by proposing and training a speculator on long-context data. The position encoding for their proposed speculator remains consistent with the target model beyond the training context, allowing it to sustain high acceptance lengths. OWL~(\cite{lee2025owl}) similarly trains a recurrent speculator for long contexts, which only conditions on the hidden state of the target model's last token, decoupling the achieved acceptance length from the prefix length. While both these methods improve acceptance length, the resulting speculator remains frozen during deployment, and thus matches the target model only on average over its training corpus. 

RAPID~(\cite{chen2025rapid}) takes a different approach: instead of training a speculator from scratch, it directly adjusts the target's distribution. RAPID runs the speculator over a retrieved context, instead of the entire input, and shifts the target's distribution toward the speculator's during verification. However, this approach alters the target's output distribution and is therefore \textit{lossy}. This is in contrast to~\ours{} and all methods discussed above, which leave the target's output distribution unchanged, and is thus orthogonal.

\subsubsection{Quantitative Comparison Against Offline Adaptation Methods}
\label{app:related-offline-quant}

As~\ours{} adapts the speculator online and preserves the target model's distribution, it is orthogonal to the prior offline methods and can in fact be applied on top of them. To showcase this, we apply~\ours{} to the released LongSpec speculators\footnote{\url{https://github.com/sail-sg/LongSpec}} on two long-context target models. The exact configuration and model pairs used for this study are listed in Table~\ref{tab:longspec-setup}. As LongSpec already reduces the training-inference mismatches described in Section~\ref{sec:diagnosis}, any further improvement can be attributed to online adaptation alone, regardless of how extensively the speculator is trained offline. Table~\ref{tab:longspec-comp} shows that~\ours{} improves the acceptance length of the LongSpec speculator by up to $13.9\%$ and $10.1\%$ on the two targets, and improves it consistently across all eight benchmarks. This reiterates our earlier finding in Section~\ref{sec:extensive} that although extensive offline training reduces the distribution mismatch between the speculator and the target, it does not eliminate the need to specialize the speculator to the request being served.

\begin{table}[H]
    \caption{Target/speculator pairs and~\ours{} configuration used for the LongSpec experiments (Table~\ref{tab:longspec-comp}). Both speculator checkpoints are officially released by LongSpec and are used as-is, without any retraining or fine-tuning. All hyperparameters are identical across the two configurations.}
    \label{tab:longspec-setup}
    \centering
    \newlength{\datacol}
    \setlength{\datacol}{\dimexpr(\textwidth - 3cm - 6\tabcolsep)/2\relax}
    \begin{tabular}{@{}>{\raggedright\arraybackslash}p{3cm} >{\raggedright\arraybackslash}p{\datacol} >{\raggedright\arraybackslash}p{\datacol}@{}}
    \toprule
    \textbf{Property} & \multicolumn{1}{c}{\textbf{longchat-7b-v1.5-32k}} & \multicolumn{1}{c}{\textbf{Llama-3-8B-Instruct-262k}} \\
    \midrule
    Target model & lmsys/longchat-7b-v1.5-32k & gradientai/Llama-3-8B-Instruct-262k \\
    Speculator   & sail/longspec-longchat-7b-v1.5-32k & sail/longspec-Llama-3-8B-Instruct-262k \\
    \midrule
    Draft tree shape        & \multicolumn{2}{c}{$[4, 16, 16, 16, 16]$ (5 draft levels $+$ 1 bonus token, AL $\leq 6$)} \\
    Sampling                & \multicolumn{2}{c}{Greedy (temperature $0$)} \\
    Max output length       & \multicolumn{2}{c}{$32$K tokens} \\
    \ours{} variant         & \multicolumn{2}{c}{Entire speculator adapted} \\
    Update stride $\mathcal{S}$       & \multicolumn{2}{c}{$10$} \\
    \bottomrule
    \end{tabular}
\end{table}

\begin{table}[H]
    \caption{Acceptance length of the released LongSpec speculators with and without~\ours{}. Although LongSpec reduces the training-inference data mismatches,~\ours{} recovers further acceptance length across all benchmarks by specializing the speculator to the request being served.}
    \label{tab:longspec-comp}
    \centering
    \setlength{\tabcolsep}{4pt}
    \begin{tabular}{l ccc ccc}
    \toprule
    \multirow{2}{*}{Benchmark}
      & \multicolumn{3}{c}{longchat-7b-v1.5-32k}
      & \multicolumn{3}{c}{Llama-3-8B-Instruct-262k} \\
    \cmidrule(lr){2-4} \cmidrule(lr){5-7}
      & LongSpec & LongSpec$+\,$\ours{} & $\Delta(\%)$ & LongSpec & LongSpec$+\,$\ours{} & $\Delta(\%)$ \\
    \midrule
    AIME 2024    & 4.7 & \textbf{4.9} & $+$5.6  & 4.8 & \textbf{5.0} & $+$5.6  \\
    AIME 2025    & 4.4 & \textbf{5.0} & $+$13.9 & 4.4 & \textbf{4.9} & $+$10.1 \\
    MATH-500     & 4.6 & \textbf{4.9} & $+$6.8  & 4.3 & \textbf{4.5} & $+$5.4  \\
    OlyBench-Math & 4.2 & \textbf{4.8} & $+$12.9 & 4.6 & \textbf{4.9} & $+$7.9  \\
    OlyBench-Phy  & 4.4 & \textbf{4.9} & $+$12.1 & 4.6 & \textbf{4.8} & $+$3.7  \\
    GPQA $\diamond$ & 4.2 & \textbf{4.7} & $+$12.3 & 4.5 & \textbf{4.8} & $+$6.6  \\
    LiveCodeBench & 4.4 & \textbf{4.7} & $+$6.7  & 4.4 & \textbf{4.5} & $+$4.2  \\
    TheoremQA  & 4.2 & \textbf{4.5} & $+$5.5  & 4.5 & \textbf{4.7} & $+$2.9  \\
    \bottomrule
    \end{tabular}
\end{table}

\clearpage
\section{Batched~\ours{} With LoRA Adapters: Implementation And Results}
\label{app:lora}

In this section, we provide the implementation details of batched~\ours{} with per-request LoRA adapters introduced in Section~\ref{sec:batching}, and report batched serving results on more benchmarks and speculators.

\subsection{Implementation Of Per-Request LoRA Adapters}

\noindent \textbf{Per-Request Adapter Pools:} \ours{} maintains the LoRA
adapters for all concurrent requests in pre-allocated pools, with one slot per request indexed by the engine's request identifier. When a request is admitted, its slot is initialized with the adapter's $B$ matrix set to zero. Thus, the adapter begins speculation from the unmodified pretrained speculator checkpoint. Upon request completion, the slot, along with its optimizer state, is released and assigned to the next request. 

\noindent \textbf{Serving With Shared Speculator Weights:} During speculation, the adapter delta for each adapted projection is computed as two batched matrix multiplications. Tokens of each request are routed only through their own adaptation slot, thereby ensuring that the adaptation state of one request is \textit{invisible} to every other request in the batch. This eliminates cross-request corruption (Section~\ref{sec:batching}) by construction. The speculator's pretrained
weights are read-only and shared across the batch, mirroring the execution model that production engines already employ for serving distinct LoRA adapters over a shared base model~(\cite{chen2024punica, sheng2024slora}).

\noindent \textbf{Per-Request Gradient Updates:} At each update,~\ours{} adapts every request in the batch independently: each request's distillation loss is computed over its own generated tokens and used as a training signal for only its own LoRA adapters. To avoid launching one backward pass per request, the per-request losses are summed and differentiated in a single batched backward pass. As no loss term touches another request's adapter, this is purely an execution optimization, and each adapter receives exactly the gradient it would have otherwise received if served in isolation. Similarly, each request carries its own optimizer state (Adam moments and step count). The strided updates and asynchronous pipelining of Appendix~\ref{app:tradeoff} carry over unchanged.

\noindent \textbf{Adapted Projections:} The set of adapted projections determines both the per-request memory footprint and the computational cost of the gradient update. Therefore, we want to select the smallest set that preserves the improvements in acceptance length. In our ablations, expanding the adapted set beyond this minimal set improves acceptance length only marginally while inflating the adaptation state and slowing every update. For DFlash, we adapt the attention query and output projections along with the MLP projections. All studies use rank $r{=}16$ adapters with a learning rate of \texttt{1e-4}.

\subsection{Batched Serving Results}

We now evaluate \ours{} under batched serving. Table~\ref{tab:batched-122b} extends the Qwen3.5-122B-A10B results of Table~\ref{tab:batched-throughput} to the remaining four benchmarks. As the baseline acceptance lengths are lower on this large
target ($1.7{-}3.1$),~\ours{} has more headroom and improves throughput by up to $2.4{\times}$.

\begin{table}[H]
\caption{Throughput (Tokens/Sec) of~\ours{} versus DFlash
(z-lab/Qwen3.5-122B-A10B-DFlash) on Qwen3.5-122B with $\mathcal{S}{=}10$ for the remaining benchmarks.~\ours{} improves throughput by up to $2.4{\times}$.}
\label{tab:batched-122b}
\centering
\setlength{\tabcolsep}{4pt}
\begin{tabular}{@{}ll r@{\,}l r@{\,}l r@{\,}l r@{\,}l@{}}
    \toprule
    Benchmark & Approach & \multicolumn{2}{c}{$BS{=}8$} & \multicolumn{2}{c}{$BS{=}16$} & \multicolumn{2}{c}{$BS{=}32$} & \multicolumn{2}{c}{$BS{=}64$} \\
    \midrule
    \multirow{2}{*}{MATH-500}
    & DFlash              & 280.3 & & 319.9 & & 435.8 & & 376.6 & \\
    & TTS~(\textit{Ours}) & \textbf{445.2} & ($1.59\times$) & \textbf{602.3} & ($1.88\times$) & \textbf{877.5} & ($2.01\times$) & \textbf{666.9} & ($1.77\times$) \\
    \midrule
    \multirow{2}{*}{TheoremQA}
    & DFlash              & 233.8 & & 289.1 & & 439.9 & & 313.0 & \\
    & TTS~(\textit{Ours}) & \textbf{511.9} & ($2.19\times$) & \textbf{661.3} & ($2.29\times$) & \textbf{987.9} & ($2.25\times$) & \textbf{663.5} & ($2.12\times$) \\
    \midrule
    \multirow{2}{*}{OlyBench-Math}
    & DFlash              & 193.9 & & 267.7 & & 392.9 & & 196.8 & \\
    & TTS~(\textit{Ours}) & \textbf{444.8} & ($2.29\times$) & \textbf{575.8} & ($2.15\times$) & \textbf{908.4} & ($2.31\times$) & \textbf{466.9} & ($2.37\times$) \\
    \midrule
    \multirow{2}{*}{OlyBench-Phy}
    & DFlash              & 197.6 & & 261.3 & & 387.7 & & 249.4 & \\
    & TTS~(\textit{Ours}) & \textbf{401.4} & ($2.03\times$) & \textbf{503.3} & ($1.93\times$) & \textbf{779.3} & ($2.01\times$) & \textbf{467.2} & ($1.87\times$) \\
    \bottomrule
\end{tabular}
\end{table}

We also evaluate on the smaller Qwen3.6-27B target. Table~\ref{tab:batched-27b} shows that~\ours{} improves throughput by $1.16{-}1.73{\times}$ across benchmarks and batch sizes.

\begin{table}[H]
\caption{Throughput (Tokens/Sec) of~\ours{} relative to DFlash
(z-lab/Qwen3.6-27B-DFlash) on Qwen3.6-27B with $\mathcal{S}{=}24$.~\ours{} consistently improves throughput by $1.16{-}1.73{\times}$ across benchmarks.}
\label{tab:batched-27b}
\centering
\setlength{\tabcolsep}{4pt}
\begin{tabular}{@{}ll r@{\,}l r@{\,}l r@{\,}l r@{\,}l@{}}
    \toprule
    Benchmark & Approach & \multicolumn{2}{c}{$BS{=}8$} & \multicolumn{2}{c}{$BS{=}16$} & \multicolumn{2}{c}{$BS{=}32$} & \multicolumn{2}{c}{$BS{=}64$} \\
    \midrule
    \multirow{2}{*}{AIME 2024}
    & DFlash              & 371.7 & & 505.2 & & 671.2 & & 807.8 & \\
    & TTS~(\textit{Ours}) & \textbf{480.4} & ($1.29\times$) & \textbf{676.3} & ($1.34\times$) & \textbf{879.9} & ($1.31\times$) & \textbf{1042.8} & ($1.29\times$) \\
    \midrule
    \multirow{2}{*}{AIME 2025}
    & DFlash              & 366.1 & & 503.3 & & 689.6 & & 827.7 & \\
    & TTS~(\textit{Ours}) & \textbf{491.0} & ($1.34\times$) & \textbf{672.0} & ($1.34\times$) & \textbf{910.4} & ($1.32\times$) & \textbf{1066.2} & ($1.29\times$) \\
    \midrule
    \multirow{2}{*}{GPQA $\diamond$}
    & DFlash              & 323.2 & & 430.2 & & 613.2 & & 693.4 & \\
    & TTS~(\textit{Ours}) & \textbf{428.7} & ($1.33\times$) & \textbf{607.5} & ($1.41\times$) & \textbf{790.8} & ($1.29\times$) & \textbf{844.8} & ($1.22\times$) \\
    \midrule
    \multirow{2}{*}{LiveCodeBench}
    & DFlash              & 329.9 & & 447.2 & & 622.3 & & 700.3 & \\
    & TTS~(\textit{Ours}) & \textbf{450.0} & ($1.36\times$) & \textbf{607.2} & ($1.36\times$) & \textbf{800.6} & ($1.29\times$) & \textbf{881.8} & ($1.26\times$) \\
    \midrule
    \multirow{2}{*}{MATH-500}
    & DFlash              & 365.8 & & 520.8 & & 650.9 & & 757.2 & \\
    & TTS~(\textit{Ours}) & \textbf{487.8} & ($1.33\times$) & \textbf{605.4} & ($1.16\times$) & \textbf{853.4} & ($1.31\times$) & \textbf{916.4} & ($1.21\times$) \\
    \midrule
    \multirow{2}{*}{TheoremQA}
    & DFlash              & 319.9 & & 427.8 & & 626.9 & & 828.2 & \\
    & TTS~(\textit{Ours}) & \textbf{553.3} & ($1.73\times$) & \textbf{614.8} & ($1.44\times$) & \textbf{862.2} & ($1.38\times$) & \textbf{1010.3} & ($1.22\times$) \\
    \midrule
    \multirow{2}{*}{OlyBench-Math}
    & DFlash              & 358.0 & & 515.7 & & 671.9 & & 717.1 & \\
    & TTS~(\textit{Ours}) & \textbf{500.5} & ($1.40\times$) & \textbf{686.0} & ($1.33\times$) & \textbf{922.4} & ($1.37\times$) & \textbf{978.4} & ($1.36\times$) \\
    \midrule
    \multirow{2}{*}{OlyBench-Phy}
    & DFlash              & 345.3 & & 464.1 & & 643.1 & & 711.8 & \\
    & TTS~(\textit{Ours}) & \textbf{481.2} & ($1.39\times$) & \textbf{651.8} & ($1.40\times$) & \textbf{866.7} & ($1.35\times$) & \textbf{939.9} & ($1.32\times$) \\
    \bottomrule
\end{tabular}
\end{table}